\documentclass[sigplan,screen,nonacm]{acmart}
\usepackage{multirow}
\usepackage{tabularray}

\usepackage[normalem]{ulem}
\useunder{\uline}{\ul}{}
\AtBeginDocument{%
  \providecommand\BibTeX{{%
    \normalfont B\kern-0.5em{\scshape i\kern-0.25em b}\kern-0.8em\TeX}}}

\begin{document}

\title{Customize StyleGAN with One Hand Sketch}

\author{Zhang Shaocong}
\email{szhang012@e.ntu.edu.sg}


\begin{abstract}
   Generating images from human sketches typically requires dedicated networks trained from scratch. In contrast, the emergence of the pre-trained Vision-Language models (e.g., CLIP) has propelled generative applications based on controlling the output imagery of existing StyleGAN models with text inputs or reference images. Parallelly, our work proposes a framework to control StyleGAN imagery with a single user sketch. In particular, we learn a conditional distribution in the latent space of a pre-trained StyleGAN model via energy-based learning and propose two novel energy functions leveraging CLIP for cross-domain semantic supervision. Once trained, our model can generate multi-modal images semantically aligned with the input sketch. Quantitative evaluations on synthesized datasets have shown that our approach improves significantly from previous methods in the one-shot regime. The superiority of our method is further underscored when experimenting with a wide range of human sketches of diverse styles and poses. Surprisingly, our models outperform the previous baseline regarding both the range of sketch inputs and image qualities despite operating with a stricter setting: with no extra training data and single sketch input.
\end{abstract}


\begin{teaserfigure}
  \includegraphics[width=\textwidth]{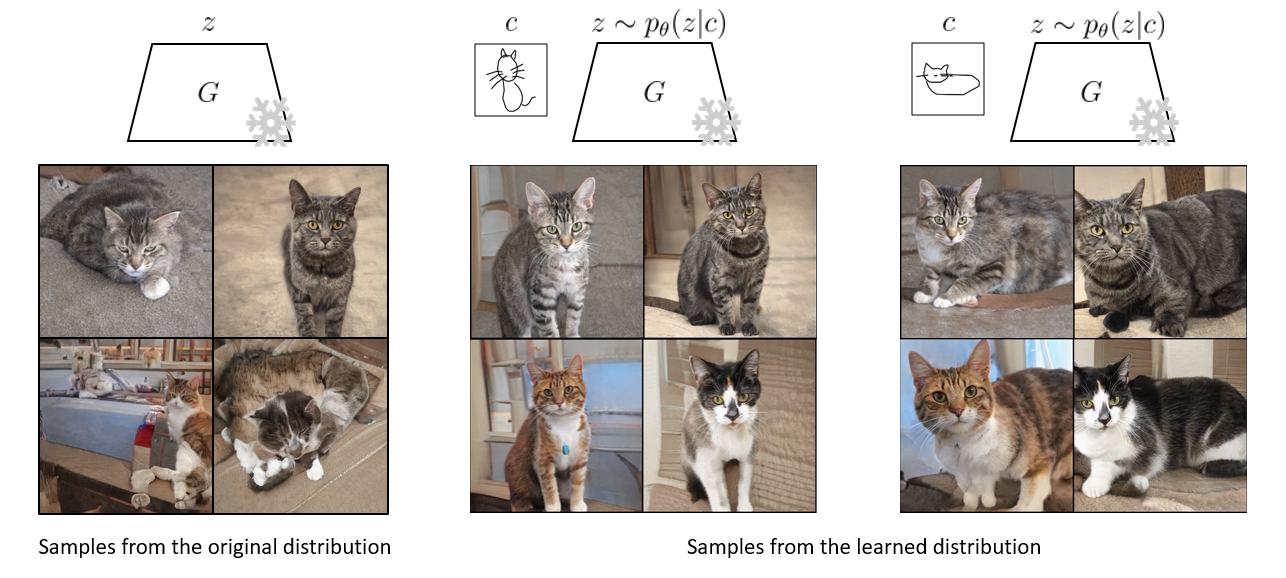}
  \caption{\textbf{ Customize StyleGAN with one human sketch. }Our method takes a single user sketch as input and controls the output imagery of a pre-trained StyleGAN by learning a conditional distribution in its latent space. Drawing noises from this learned distribution and feeding into the frozen source model will produce multi-modal image samples semantically
aligned to the sketch.  }
  \label{fig:teaser}
\end{teaserfigure}

\maketitle

\section{Introduction}
Generative Adversarial Networks (GANs) \cite{goodfellow2014generative} have opened up exciting possibilities in content generation owing to their remarkable ability to produce photorealistic images. In particular, StyleGAN \cite{karras2019style,karras2020analyzing} is further revealed to possess inherent disentangled properties \cite{harkonen2020ganspace}. Aligning perfectly with a crucial requirement in real-world graphic applications: an intuitive interface to apply users' creative intents, this discovery has inspired a wave of controllable synthesis methods rooted in manipulating StyleGAN-generated imagery \cite{shen2020interpreting,wu2021stylespace,abdal2021styleflow}. The paradigm shift is further propelled by the fusion of StyleGAN with Contrastive Language-Image Pre-training (CLIP) \cite{patashnik2021styleclip}, enabling users to exert control over model-generated imagery through textual descriptions alone.

As one observes this remarkable progress, a tantalizing question arises: Can we achieve a similar way of control over pre-trained StyleGAN's imagery by user sketch? Despite having been extensively explored, sketch-guided generation under current methodologies often necessitate dedicated frameworks and datasets. In contrast, we aspire to align with the contemporary trend of harnessing the abundant domain knowledge inherent in pre-trained generative models, thus circumventing the redundant reinvention of the wheel. While a similar concept has been explored by \cite{syog}, our approach leverages CLIP models \cite{clip} for supervision, eliminating the need for adversarial training. This feat is achievable through CLIP-space's extraordinary capability to encapsulate high level semantic concepts, bridging domains across not only vision and language but diverse visual representational styles as well.

In this paper, we introduce a CLIP-based approach designed to repurpose existing StyleGAN models to generate multi-modal, photorealistic images aligned with a specific user sketch. Notably, we are the first to harness CLIP for semantic guidance in sketch-based synthesis. Our method also draws inspiration from other concurrent research, particularly PromptGen \cite{wu2022generative}, which reformulates the task of controllable synthesis as learning a conditional distribution within the latent space of a frozen generative network via energy-based modeling (EBM). We extend this scheme for sketch-guided synthesis by designing novel energy functions leveraging CLIP's cross-domain semantic knowledge. Furthermore, we acknowledge that sketch-guided synthesis is inherently multi-modal, as reference sketches should convey object-level concepts without specifying colours, textures, or intricate details. To address this, we rely on StyleGAN's inherent disentangled structure and employ style-mixing techniques introduced in \cite{karras2020analyzing}, supervising the learning only along content-related directions in CLIP space.

We conduct quantitative comparisons against existing methods with the synthesized dataset published in \cite{syog} and then proceed to apply our approach to real hand-drawn sketches from QuickDraw \cite{cheema2012quickdraw} and Sketchy \cite{sangkloy2016sketchy}, which presents a wide range of diverse sketch styles and poses. Unlike previous methods, our model requires no additional datasets to train and only takes one user sketch and a neutral text description (e.g., 'cat') as inputs. Nevertheless, it still outperforms the baseline quantitatively in the one-shot regime. Despite our more stringent training setting, The advantage is even more remarkable with real sketches. Despite our more stringent one-shot setting, our method not only significantly improves upon previous works in terms of output qualities but also excels in its ability to adapt to various sketch styles and poses. As demonstrated in Figure \ref{fig:teaser}, despite the sketches being overly abstract and distorted, the output samples are photorealistic and successfully aligned with the main concepts in the sketches (e.g., pose, position). Finally, we present several examples to illustrate that our method could smoothly integrate with other StyleGAN-based manipulations like latent space editing and natural image inversion.

\section{Related Work}
\textbf{Sketch-guided deep image synthesis. }
Deep solutions of sketch-based synthesis typically approach the task as image-to-image translation. Scribbler \cite{sangkloy2017scribbler} and SketchyGAN \cite{chen2018sketchygan} are among the first works to propose end-to-end solutions explicitly targeting human sketch inputs. However, these data-oriented approaches require paired sketch-image datasets for training, the collecting of which requires substantial human effort. Further, human sketches' varying styles, distortions, and abstraction levels amplify the challenge for such datasets to be distributionally representative. An alternative approach is to explore unsupervised or self-supervised learning \cite{liu2020unsupervised,liu2021self,cheng2023adaptively}.
Nevertheless, these methods still necessitate training dedicated generative models from scratch. In contrast, several works \cite{syog,koley2023picture} attempt to tackle this task from a transfer learning \cite{wang2018transferring} perspective. GANSketch \cite{syog} introduced a novel cross-domain adversarial loss to rewrite pre-trained GAN weights, thus generating multi-modal images aligned with the reference sketches. \cite{koley2023picture} proposed a novel decoupled training paradigm, splitting the sketch-to-image (S2I) task into two sub-stages: sketch-to-image latent mapping and image generation, while using a pre-trained GAN model to handle the generative stage. Nonetheless, the mapping stage still requires paired data to train. Our work follows a similar intuition of leveraging pre-trained generative models. However, we distinguish ourselves from previous works by obviating the need for adversarial training or extra training datasets.

\textbf{Controllable GANs. }
Early approaches of Controllable synthesis in GANs involved training dedicated networks that took explicit controlling parameters as input \cite{li2019controllable, shoshan2021gan}. 
Recent advancements have shifted towards leveraging the intrinsically disentangled latent space \cite{wu2021stylespace} of pre-trained StyleGAN models. Instead of explicit parameters, controllable generation is achieved by manipulating sampled latents along pre-discovered directions within the latent space \cite{abdal2021styleflow,shen2020interpreting}. StyleCLIP \cite{patashnik2021styleclip} is the pioneering work to pair this approach with CLIP, allowing text-guided control over StyleGAN imagery. Further extending the idea of CLIP-based supervision, recent methods \cite{gal2022stylegan,zhu2021mind,kwon2023one} have applied these techniques to few-shot GAN adaptation, showcasing the possibility to synthesize images with desired attributes by finetuning GAN with only textual descriptions or reference images. However, they rely on ad-hoc regularization methods like identity loss \cite{richardson2021encoding} or partial model freezing to enforce in-domain constraints, which either potentially limits the scope of target imagery or leads to outputs not preserving necessary features from the original domain. As observed in \cite{wu2022generative}, when conditioned on an in-domain attribute 'baby (-like),' existing methods like StyleCLIP and StyleGAN-NADA \cite{gal2022stylegan} either fail to give condition-compliant outputs or produce unrealistic-looking ones. 
Conversely, alternative approaches \cite{plug,nie2021controllable,wu2022generative} propose to directly learn conditional distributions of GANs' output imagery as energy-based models (EBMs) \cite{lecun2006tutorial}. In this way, in-domain constraints are explicitly modelled as distribution priors. Moreover, EBMs associate data probability with energy functions; the latter can be flexibly designed to accommodate various controlling conditions.PromptGen \cite{wu2022generative} demonstrates several such conditions, performing image synthesis with specific poses or textual descriptions leveraging off-the-shelf models like inverse graphics models and CLIP.
Our work extends this framework by introducing novel energy functions that leverage CLIP-space embeddings to quantify cross-domain semantic distances. This enhancement enables our model to sample photorealistic images from the source GAN's image space while conforming to the semantic concepts within input sketches.

\textbf{CLIP for semantic supervision}
The CLIP model \cite{clip}, trained on an extensive corpus of paired image-text data, has demonstrated its capacity to encompass rich semantic knowledge within its joint vision-language embedding space. This advancement has ushered in opportunities for image-free supervision, catalyzing innovative approaches across image generation \cite{nichol2021glide}, manipulation \cite{patashnik2021styleclip, Brooks_2023_CVPR}, and image-to-sketch translation \cite{chan2022learning, Lee_2023_CVPR} tasks.
StyleCLIP \cite{patashnik2021styleclip} is among the first to harness CLIP for semantic supervision, allowing text-guided manipulation of StyleGAN imagery by directly minimizing CLIP scores between images and input texts. StyleGAN-NADA introduces an improved directional CLIP loss for text-guided domain adaption of GAN models. In parallel, \cite{chan2022learning} investigates an Image-to-Sketch training scheme leveraging CLIP scores as one of several supervisional constraints. Their findings highlight the efficacy of minimizing CLIP-space distances between synthesized line drawings and input photos as a potent method for cross-domain semantic supervision.
These pioneering works underscore CLIP's exceptional representational capacity, transcending different representational domains. Building upon this insight, we introduce a novel approach to leveraging CLIP to bridge the semantic gap between images and sketches.

\begin{figure*}[h!t]
  \centering
  \includegraphics[width=\linewidth]{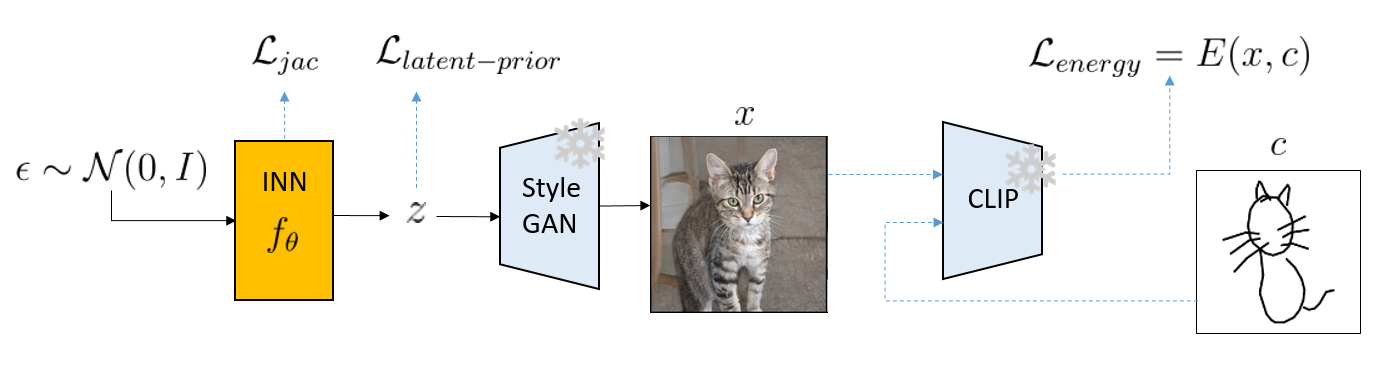}
  \caption{\textbf{Training procedure.} We train an INN to approximate the target conditional distribution in the input latent space of a frozen StyleGAN. In each forward pass, we sample $\epsilon \sim \mathcal{N}(0,I)$, which is then mapped to input noise $z$ by INN and translated to image through the frozen generator. $f_\theta$ is optimized with three loss component: $\mathcal{L}_{jac}$ the Jacobian determinant loss for INN, $\mathcal{L}_{z-prior}$ the latent prior loss and $\mathcal{L}_{energy}$ the loss defined by a custom scalar-value energy function. We leverage a frozen CLIP to implement novel value functions to measure the semantic similarity between image samples and reference sketches. }
\label{fig:network}
\end{figure*}

\section{Methods}

Given a source StyleGAN model, we aim to re-purpose it to produce samples from the model’s image domain that are semantically aligned with a reference sketch.

In the following, we describe our approach starting by introducing the preliminaries of the latent-space EBM method for controllable image synthesis (Section \ref{sec:preliminaries}). The main novelty of our research is the two CLIP-based novel energy functions we designed for the EBM framework, which serve to encourage semantic similarity between output samples and reference sketches (Section \ref{sec:dir}, \ref{sec:clipnce}). Additionally, we address the inherent multi-modality of sketch-guided synthesis harnessing StyleGAN’s intrinsic style-mixing techniques (Section \ref{sec:stylemix}).

\subsection{Preliminaries}
\label{sec:preliminaries}
Several works \cite{plug,nie2021controllable,wu2022generative} formulate the tasks of controlling GAN imagery as learning conditional distributions in the pre-trained models' latent space as EBMs \cite{lecun2006tutorial}. We briefly explain this method following similar notations as \cite{wu2022generative}.

Given a guiding sketch $c \in C$, and image samples $x \in X$, where X is the image space of a source model $G$ and $C$ the sketch domain, the core concept of EBM is to associate each possible configuration of $x$ and $c$ with an unnormalized scalar measurement through $E(x,c):X \times C \rightarrow R$ , denoted as the energy function.

Using Gibbs distribution, the conditional density of $c$ given $x$ can be formulated as the following EBM:

\begin{equation} \label{eq1}
p(c|x) =\frac{e^{-\lambda E(x,c)}}{Z_c}, Z_c = \int_{c'\in C} e^{-\lambda E(x,c')}
\end{equation}

Here $\lambda$ is a weight hyperparameter, and $Z_c$ is a normalizing constant integrated over $C$, denoted as the partition function, which is intractable as $C$ represents the hypothetical space of all possible sketches. However, since $C$ and $X$ are both fixed, we note that $Z_c$ and all following derivations of partition functions are constants not affecting the final training objective.

We then rewrite the conditional density over image space $X$, combining Bayes’s rule and equation \ref{eq1}:

\begin{equation} \label{eq2}
\begin{split}
p(x|c) \propto p_x(x)p(c|x)= \frac{p_x(x)e^{-\lambda E(x,c)}}{Zx} ,\\
Z_x = \int_{x'\in X} p_x(x')e^{-\lambda E(x',c)} 
\end{split}
\end{equation}

The image prior $p_x(x)$ is defined by sampling a random noise $z\sim \mathcal{N}(0,I)$ and then translating it to image space through the frozen source GAN, $x=G(z)$. The above EBM is thus reformulated in the latent space:

\begin{equation} \label{eq3}
P(z|c) = \frac{ P_z(z)e^{-\lambda E(G(z),c)}}{Z_z}, Z_z = \int_{z'} p_z(z')e^{-\lambda E(G(z'),c)}
\end{equation}

$p(z|c)$ can be trained with a parameterized network $p_{\theta}(z)$ by minimizing the KL divergence $\mathbb{D}_{KL} (p_\theta(z) \lVert p(z|c))$. PromptGen proposes to model $p_{\theta}(z)$ as an invertible bijection $f_\theta$ \cite{inn}, which maps a Gaussian noise to the latent space, $z = f_\theta(\epsilon), \epsilon \sim \mathcal{N}(0,I)$.

With $f_\theta$’s invertible properties, the probability density of z has a closed-form formula:

\begin{equation} \label{eq4}
logp_\theta(z)=log\mathcal{N}(\epsilon | 0,I)+log|det(\frac{\partial f_\theta}{\partial \epsilon})|
\end{equation}

Combing with equation \ref{eq3}, the full objective used to train $f_\theta$  is:

\begin{align*}\label{eq5}
&\operatorname*{argmin}_\theta E_{\epsilon \sim \mathcal{N}(0,I),z=f_\theta(\epsilon),x=G(z)}[\mathcal{L}_{jac}+\mathcal{L}_{z-prior}+\mathcal{L}_{energy}], \\
&\mathcal{L}_{jac}=-log|det(\frac{\partial f_\theta}{\partial \epsilon})|,\\ 
&\mathcal{L}_{z-prior}=-logP_z(z), \\
&\mathcal{L}_{energy}=E(x,c)
\end{align*}

We illustrate the training procedure in Figure \ref{fig:network}.
The first two loss components are the log-determinant of the Jacobian for $f_\theta$ and the latent prior loss. The third term $E(x,c)$ is the energy function, which by theory could be any function that associates a scalar value to each generated sample $x$ given reference sketch $c$, where smaller values are assigned to the preferred data samples, in our case, samples semantically similar to $c$. The following two sections discuss how to implement such a cross-domain semantic similarity measurement as the energy function leveraging CLIP.

\subsection{Cross-domain directional CLIP energy}
\label{sec:dir}
A naive way to leverage CLIP for semantic supervision is to minimize the CLIP-space cosine distance between an image sample $x$ and the reference sketch $c$ \cite{patashnik2021styleclip}:

\[
E_{\text{global}}(x,c) = 1 - cos(F_I(x), F_I(c))
\]
where $cos(.)$ stands for cosine similarity, and $F_I$ is CLIP’s image encoder.

However, owing to the non-trivial domain gaps between images and sketches, optimizing CLIP scores directly leads to content-irrelevant visual features of the sketch domain leaking into generated images. Inspired by StyleGAN-NADA \cite{gal2022stylegan}, we instead define a pair of source anchors in the CLIP space, one in each representational domain. This source pair is introduced by first choosing an image anchor $x_o \in X$, and then employing an off-the-shelf Image-to-Sketch network $H$ (Photosketch\cite{li2019photo}) to generate its sketch-domain counterpart, $c_o = H(x_o)$. The cross-domain CLIP directional energy can then be defined as:
\[\Delta C = F_I(c) - F_I(c_o)\]
\[\Delta X = F_I(x) - F_I(x_o)\]
\[E_{Dir}(x,c) = 1 - cos(\Delta C, \Delta X)\]

For each $x$ sampled from the target distribution, we align the CLIP-space direction between image anchor $x_o$ and $x$ to the direction between sketch anchor $c_o$ and target sketch $c$. We rely on the assumption that the CLIP-space distance of two data points within the same representational domain (e.g., $\Delta C$) depicts only their content-related semantic difference while excluding common representational features of that artistic domain. Plus $x_o$ and $c_o$ being cross-domain semantic equivalents, by aligning $\Delta X$ to $\Delta C$, one essentially encourages $x$ to differ from a source concept (defined by $(x_o,c_o)$) along the same semantic direction from that source concept to the target concept of the reference sketch.
 
 Similar to using category names as source anchors in \cite{gal2022stylegan}, one should ideally choose an image source anchor to represent a neutral mode in the data space. However, choices of such source points in the image domain are not immediately obvious. We empirically use $G(w_{avg})$ ($w_{avg}$ being the average latent of the pre-trained StyleGAN model), which we find to give decent performance.

\begin{figure}[t]
  \centering
  \includegraphics[width=\linewidth]{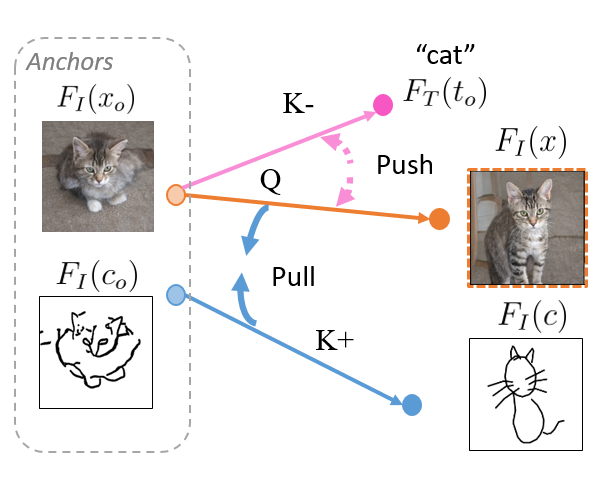}
  \caption{\textbf{Illustration of CLIP-NCE energy.} The contrastive loss is computed in the CLIP space, where the query $Q$ (orange arrow) is defined by the directions from the embedding of the source image anchor to that of the generated samples. The positive sample $K+$ (blue arrow) is the direction from the sketch anchor embedding to that of the reference sketch. We additionally use a negative sample $K-$ (pink arrow), which is the direction from the source image anchor to the embeddings of a neutral category-level description $t_o$ (e.g.,' cat'), from which we wish to push $Q$ away. $F_I$ represents the CLIP image encoder and $F_T$ the CLIP text encoder. The pair of anchors are chosen as such: $x_o$ is a hand-selected sample from the source GAN (empirically, we use $G(w_{avg})$, the image sample generated from the average $W$ space latent), while $c_o$ its sketch counterpart generated from an off-the-shelf I2S network.   }
\label{fig:nce}
\end{figure}

\subsection{Cross-domain CLIP-NCE energy}
\label{sec:clipnce}
We find that using directional energy alone is prone to produce artefacts, which we will discuss in later sections. Therefore, we propose a second supervision scheme using InfoNCE-like contrastive learning \cite{infonce} following similar notations as \cite{cfclip}. In contrastive learning, positive samples are pulled closer to a query vector $Q$, while negative samples are simultaneously pushed farther away. Following this practice, we first define the query $Q$, the negative sample $K-$ and the positive sample $K+$ as follow:

$$
Q = \Delta X = F_I(x) - F_I(x_o)
$$

$$
K+ = \Delta C = F_I(c) - F_I(c_o)
$$

$$
K- = F_T(t_{o}) - F_I(x_o)
$$
The first two terms are identical to the directional energy formulas. $\Delta X$, the direction from the source image anchor to the target image sample, is defined as $Q$, while $\Delta C$ is the positive sample. Additionally introduced is the negative sample $K-$. Here $t_{o}$ is a supplied category-level description (e.g., ’cat’) and $F_T$ is CLIP’s text encoder. We utilize the textual embedding of the category label to represent a neutral mode in the source image space from which we aim to push the generated samples away. We find this also empirically mitigates modification over entity-unrelated features (e.g., background) and regularizes the samples to be more realistic-looking. 
The full energy function is defined as follows using the same formula of InfoNCE loss, where $r$ is a temperature hyperparameter:
$$
E_{NCE}(x,c) = 1 - \frac{e^{(Q \cdot K^+/r)}}{e^{(Q \cdot K^+/r)}+e^{(Q \cdot K^-/r)}}
$$

\subsection{style-mixing}
\label{sec:stylemix}
Sketch-conditioned synthesis intrinsically calls for multi-modality: the sketch does not dictate color, texture or any other detailed visual features of the target outputs. Rather, we identify it to express a high-level semantic concept and should only control the output’s rough pose, shape and position. This requirement naturally aligns with StyleGAN's disentangled structure. Specifically, StyleGAN maps an input noise $z$ to an intermediate 512-dimensional latent $w$, which then goes through a different affine transformation to generate style parameters for each synthesis network layer. Preliminary researches have demonstrated that these style parameters at different layers control different levels of detail in the output images, which is roughly divided into three groups: coarse, medium and fine. A common method to achieve multi-modal generation is thus to use a second $w'$ to control selected layers of the synthesis network \cite{richardson2021encoding}. We adopt this style-mixing technique \cite{karras2020analyzing} in training when synthesizing images to be evaluated by the energy function. Specifically, in the computation of $\Delta X = x - x_o$, we supply to the medium-to-fine layers (5-last) of the synthesis network a random $w’$ sampled from source model's $W$ space both when generating the image sample $x$ and the image anchor $x_o$. That is, for each $\Delta X$, the pair of $x$ and $x_o$ is generated with the same detailed features. Our parameterized distribution is thus restraint to only learn coarse-level concepts without undermining the source model’s style diversity. The same procedure is also employed in reference time to generate multi-modal outputs with diverse visual details.

\begin{figure*}[t!h]
  \centering
  \includegraphics[width=\linewidth]{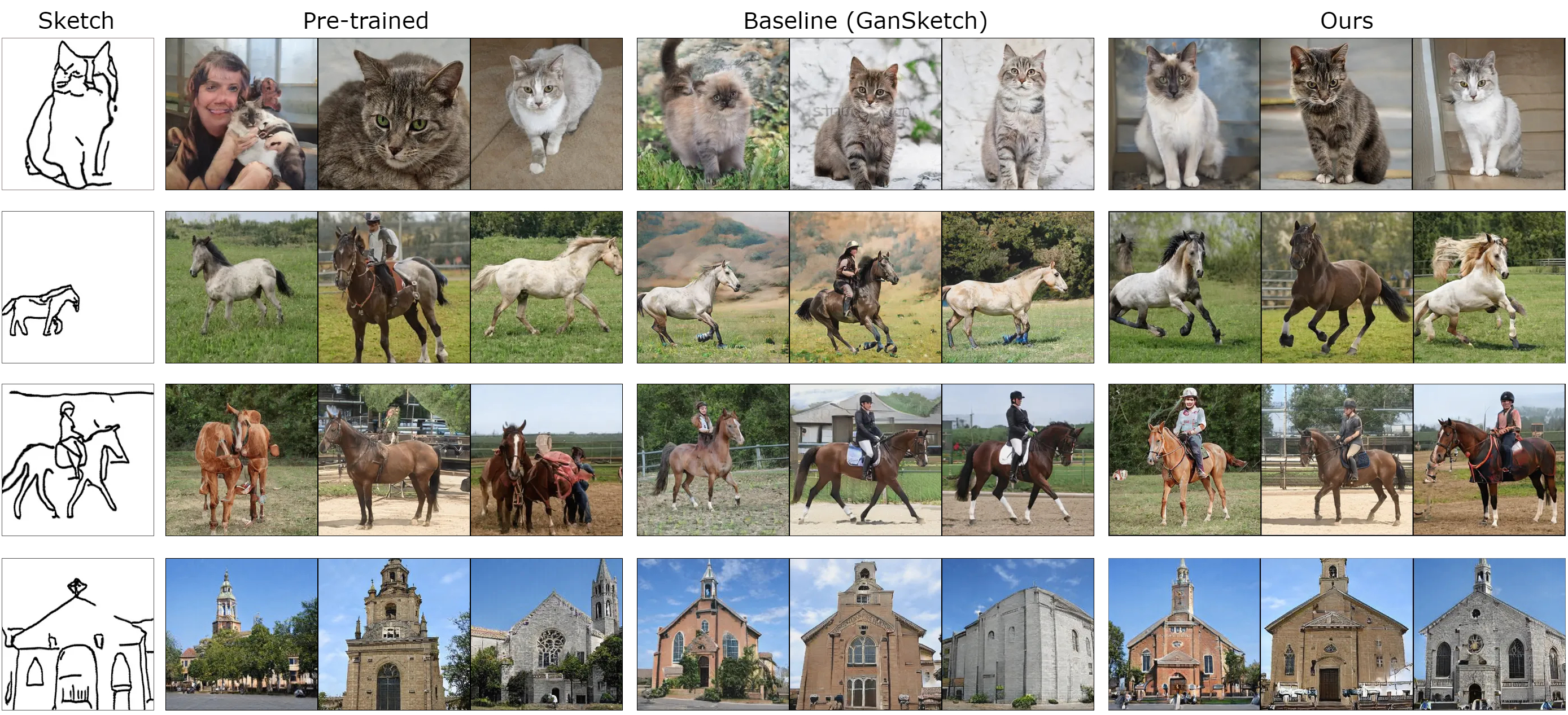}
  \caption{\textbf{Qualitative comparisons on synthetic sketches. } We compare samples generated from the original pre-trained models, the GANSketch \cite{syog} models and our models on four synthetic test cases. GANSketch samples are generated with their best model variants trained with 5 to 30 sketch inputs. Our models are each trained with one input sketch and we show the samples of the best model variants as reported in Table \ref{tablefid}. All samples are generated with the same set of input noise and truncation $\Phi=0.5$, following StyleGAN2 \cite{karras2020analyzing}.    }
\label{fig:photosketch}
\end{figure*}

\begin{table*}[]
\begin{tblr}{lllllll}
\hline
\SetCell[r=2]{c}{Family} &
\SetCell[r=2]{c}{Model \\ variant} &
\SetCell[r=2]{c}{No. \\ Sketches} & &&FID $\downarrow$ & \\ \cline{4-7} 
  &&& standing cat &
  horse rider &
  horse on side &
  garbled church \\ \hline
Pre-trained &
  Original &
  N.A &
  {\color[HTML]{9B9B9B} 58.71} &
  {\color[HTML]{9B9B9B} 50.43} &
  {\color[HTML]{9B9B9B} 42.24} &
  {\color[HTML]{9B9B9B} 32.64} \\ \hline
 \SetCell[r=2]{c}{GanSketch}&
  Best* &
  30(5) &
  {\color[HTML]{656565} {\ul 31.20}} &
  {\color[HTML]{656565} 19.94} &
  {\color[HTML]{656565} {\ul 29.62}} &
  {\color[HTML]{656565} {\ul 16.70}} \\
 \cline[dashed]{2-7} 
 &
  1-example &
  1 &
  {\color[HTML]{000000} 44.68} &
  {\color[HTML]{000000} 29.25} &
  {\color[HTML]{000000} 41.50} &
  {\color[HTML]{000000} 26.88} \\ \hline
 \SetCell[r=3]{c}{Ours}&
  NCE &
  1 &
  {\color[HTML]{000000} 35.05} &
  {\color[HTML]{000000} 23.21} &
  {\color[HTML]{000000} \textbf{38.41}} &
  {\color[HTML]{000000} 22.53} \\ \cline[dashed]{2-7} 
 &
  Dir &
  1 &
  {\color[HTML]{000000} 38.65} &
  {\color[HTML]{000000} 33.31} &
  {\color[HTML]{000000} 44.89} &
  {\color[HTML]{000000} \textbf{20.49}} \\
 &
  NCE+aug &
  1 &
  {\color[HTML]{000000} \textbf{34.14}} &
  {\color[HTML]{000000} {\ul \textbf{18.78}}} &
  {\color[HTML]{000000} 40.18} &
  {\color[HTML]{000000} 38.06} \\ \hline
\end{tblr}
\caption{\textbf{Quantitative analysis. } We report the Frechet Inception Distance (FID) of the pre-trained models, the baseline (GANSketch) and our methods on four synthetic evaluation cases. $\downarrow$ indicates that lower scores are better. We took three sets of GANSketch scores directly from their report. Best* are their best scores among all model variants trained with 30 sketches (5 in the case of standing cat). '1-example' denotes their one-shot results. We test on three variants of our method: 'NCE', our base variant trained with CLIP-NCE energy; 'Dir', the alternative of training with directional CLIP energy, and 'NCE+aug', which additionally use augmentation with CLIP-NCE. A fair comparison under one shot constraint is made only against GANSketch's '1-example' models and the best scores are highlighted in \textbf{black}. The best FID regardless of settings are marked with \underline{underline}.  }
\label{tablefid}
\end{table*}
\section{Experiments}

\begin{figure*}[th]
  \centering
  \includegraphics[width=\linewidth]{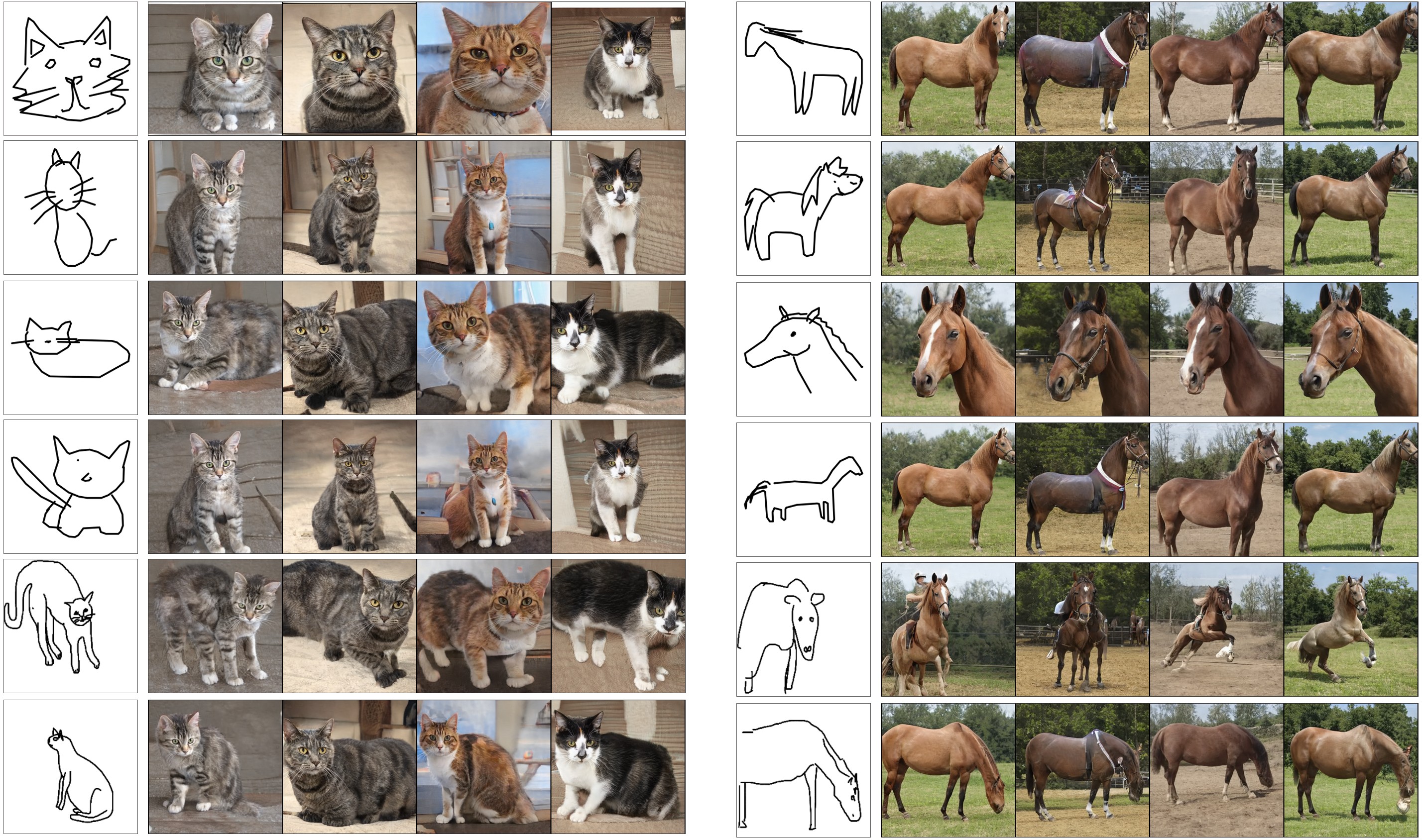}
  \caption{\textbf{Qualitative results on real-life sketches. }The first columns show the one-shot input sketches selected from the QuickDraw (row 1-4) and the Sketchy (row 5-6) datasets. The following columns are the uncurated images sampled from the model trained with that input sketch. All samples are generated with truncation $\phi=0.5$.   }
\label{fig:handsketch}
\end{figure*}

\textbf{Datasets. }
We use GANSketch’s published dataset. The dataset provides 4 test cases for quantitative evaluation: 1) standing cat 2) horse on a side 3) horse rider 4) garbled church. Each case contains 30 Photosketch-generated \cite{li2019photo} sketches and 2500 real images picked from the LSUN datasets \cite{yu2015lsun} to present the target distribution. To test in real world scenarios, GANSketch datasets also includes sketches picked from QuickDraw \cite{cheema2012quickdraw}. As target image distributions for human sketches are hard to define, the results are only evaluated qualitatively. We additionally select real sketches from the Sketchy dataset \cite{sangkloy2016sketchy} to test our method with a broader range of sketch styles and object poses.

\textbf{Baselines. }
We compare our method to the pioneer work GANSketch. For quantitative cases, we compute the Frechet Inception Distance (FID) \cite{fid} scores between the generated images and the evaluation set, then select the best FID among all training iterations.

The quantitative results are reported in Table \ref{tablefid}. The GANSketch scores are taken directly from their report. Note that all our models are trained with single reference sketch while the GANSketch best cases require either 30 (horse rider, horse on side, garbled church) or 5 (standing cat) input sketches. We observe that despite our scores do not always outperform their best model variants, it improves significantly from their one-shot results. We further showcase in Figure \ref{fig:photosketch} that our output images are of comparable qualities.

\textbf{Ablation. }
We first conduct ablation experiments quantitatively on photosketch cases and report the FID scores in Table \ref{tablefid}. As photosketch-generated sketches are still different from human sketches, we proceed to investigate the individual effects of augmentation and the two types of energy functions with real sketches.

\begin{figure}[h]
  \centering
  \includegraphics[width=\linewidth]{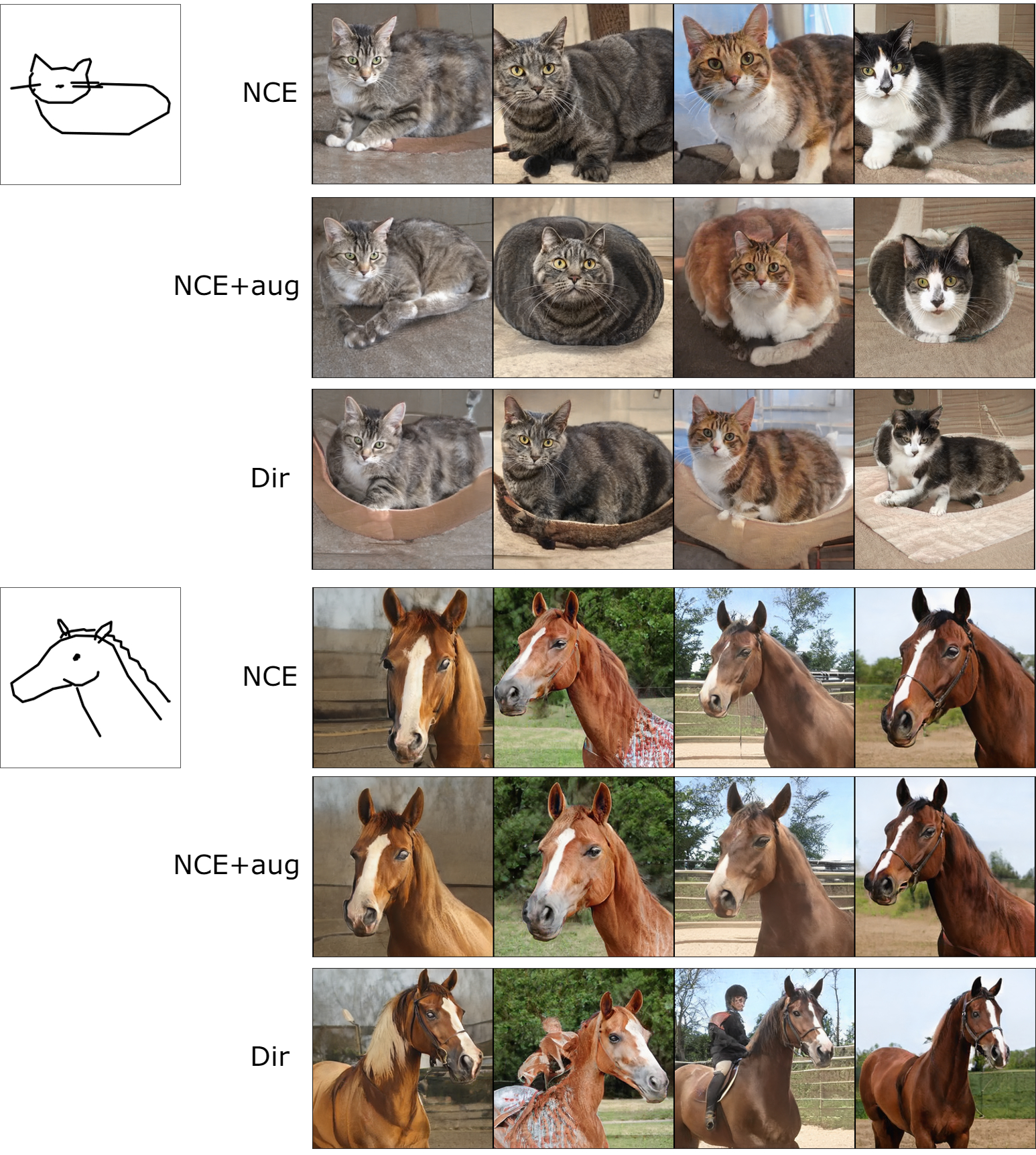}
  \caption{\textbf{Qualitative ablation studies on real-life sketches. } We experiment on human sketches to study the effects of using augmentation (models marked with '+aug') and the two types of energy functions (CLIP-NCE energy, denoted as 'NCE', and directional CLIP energy, denoted as 'Dir'). We find training without augmentation and with CLIP-NCE energy function generates the most realistic-looking samples. }
\label{fig:ablation}
\end{figure}

\textbf{Augmentation. }
Same as GANSKetch, we adopt the translation policy of differentiable augmentation \cite{diffaug}. As shown in Table \ref{tablefid}, we find that augmentation do not necessarily improve the performance on synthesized sketches. Particularly, cases with translation-invariant concepts (standing cat, horse rider) benefits from augmentation, while others (horse on side, garbled church) do not, which agrees with the translation-based scheme we choose. When testing on human sketches, we observe that using augmentation in some cases is prone to push certain target semantics to extreme (e.g, a very round cat), leading to sketch-conforming but unrealistic-looking results, as demonstrated by the 'lying cat' case in Figure \ref{fig:ablation}. On the other hand, this adverse effect is not observed when the input sketch is relatively accurate in shape (the 'horse head' case). As distortions and overly abstract depictions are common in real-life sketches (as can be observed from the input sketches in Figure \ref{fig:handsketch}), we do not apply augmentation when training on all real-sketch test cases in Figure \ref{fig:handsketch}.

\begin{figure}[h]
  \centering
  \includegraphics[width=\linewidth]{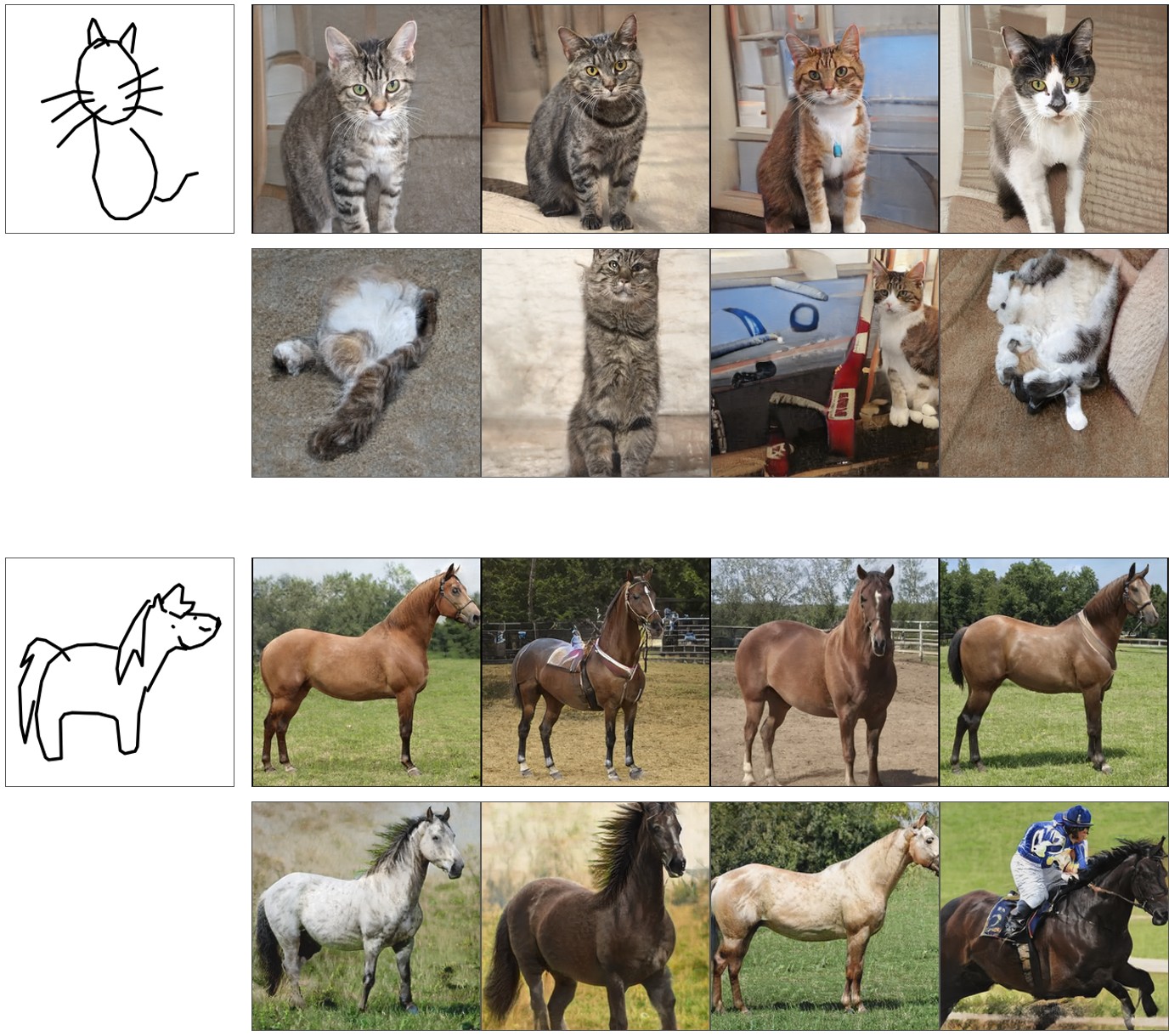}
  \caption{\textbf{Qualitative comparison with real sketches.  }The first row in each case is our result and the second row is the result generated from GANSketch's published models (one-shot variants). Both ours and GANSketch models are trained with the same sketch input, as displayed in the first column. All samples are generated with truncation $\phi=0.5$ and the same set of input noise.  }
\label{fig:compare}
\end{figure}

\begin{figure}[h]
  \centering
  \includegraphics[width=\linewidth]{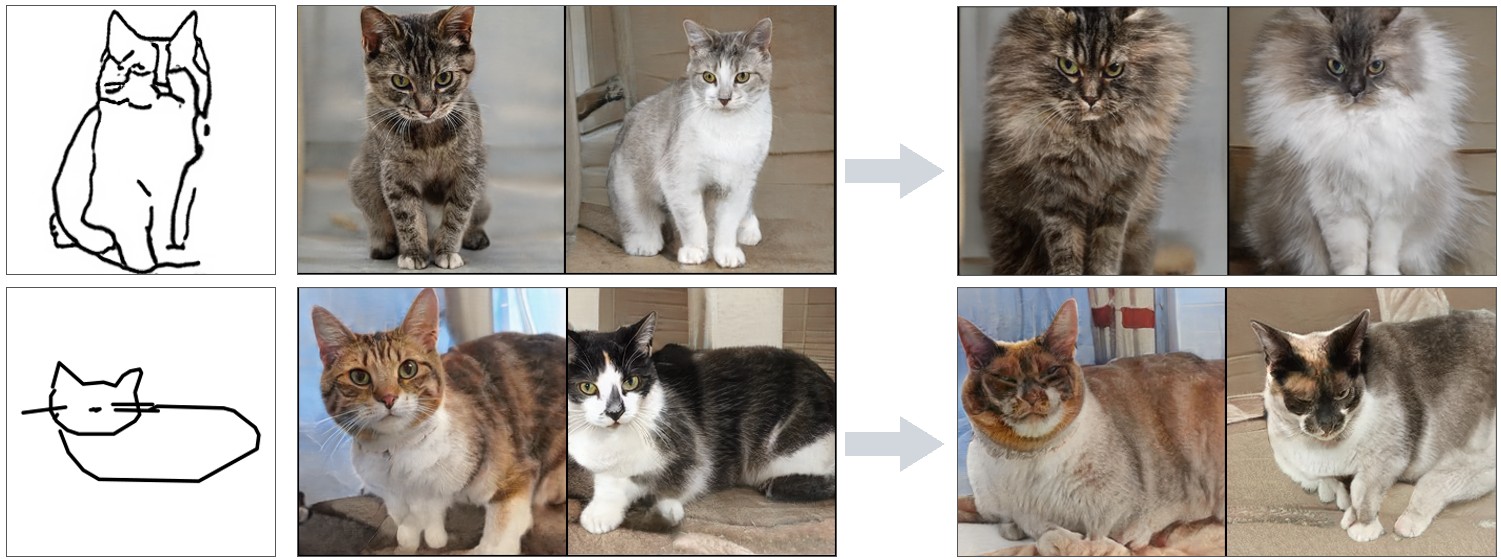}
  \caption{\textbf{Latent editing on controlled outputs.  } We experiment with applying the edit directions reported in \cite{harkonen2020ganspace} which are discovered on the source model to our controlled models. As our method do not modify source GAN weights, we find the effects of pre-discovered editing directions remain unchanged for $z$ sampled from our learned distribution.}
\label{fig:latentedit}

\end{figure}

\begin{figure}[h]
  \centering
  \includegraphics[width=\linewidth]{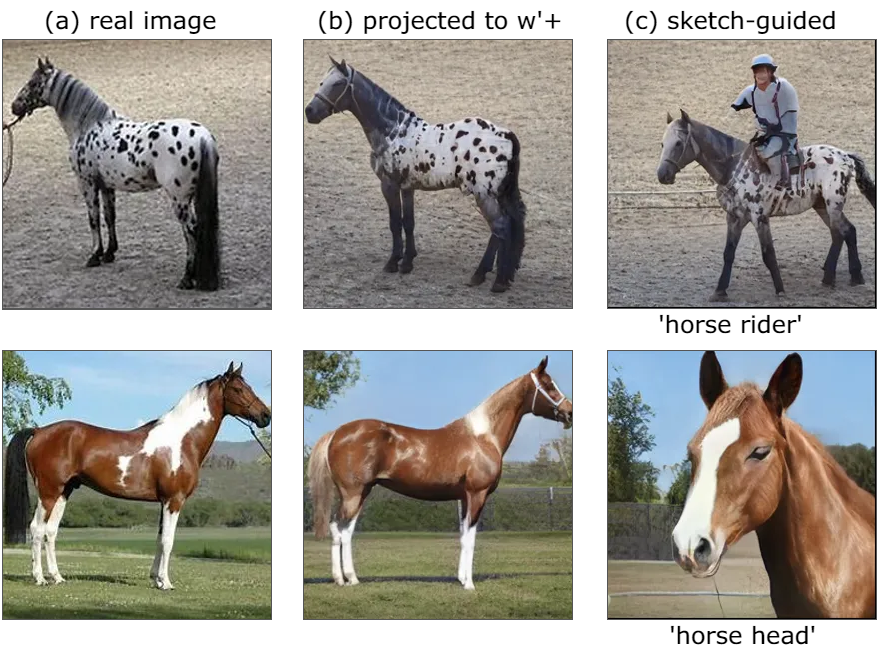}
  \caption{\textbf{Real image editing. } Given a real image as (a), we could invert it as a latent $w'+$ (b) and perform style-mixing with a randomly sampled $z$ from our sketch-conditioned distribution. The procedure of style-mixing is illustrated in Section \ref{sec:stylemix}. The result is an edited version of the natural image (c) conforming to the initial reference sketch.}
\label{fig:realinvert}
\end{figure}

\begin{figure}[h]
  \centering
  \includegraphics[width=\linewidth]{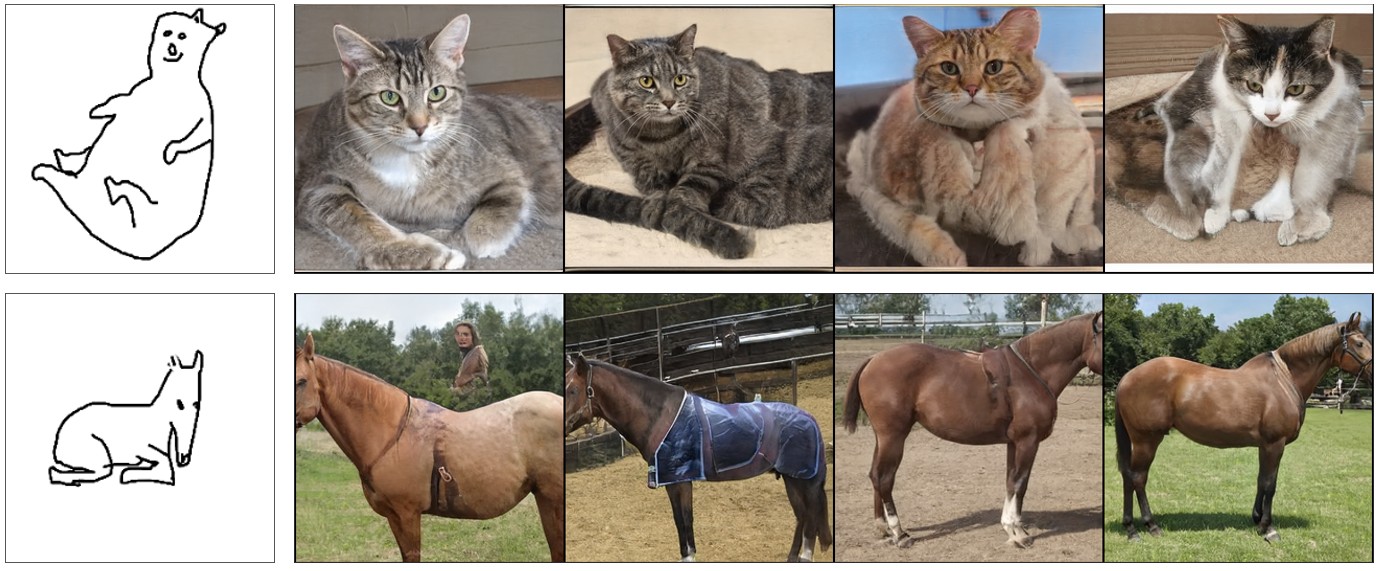}
  \caption{\textbf{Fail cases. } Our method tends to produce overall nonconforming outputs when training on sketches that represents rare modes in the source model, despite being able to capture certain semantic attributes correctly.}
\label{fig:fail}
\end{figure}

\textbf{CLIP-NCE energy versus Directional CLIP energy. }
As reported in Table \ref{tablefid}, CLIP-NCE energy outperforms directional CLIP energy in three out of four synthesized cases. We theorize the reason behind to be CLIP-NCE energy’s extra regularizing effect through contrastively learning against a neutral mode defined by the category-level text, which helps confining the learned semantics to center around the object class. This extra regularization is unessential for synthesized sketches, which possess precise outlines. In contrast, real-life sketches often contain ambiguous and distorted shapes, in which case using directional energy is prone to produce object-unrelated artefacts, as illustrated in \ref{fig:ablation}. For example, using $E_{Dir}$ on the ‘horse head’ sketch leads to manifesting the wavy line depicting mane as human-shape artefacts in output samples. A interesting comparison can be furthered made between the 'NCE+aug' outputs with the directional energy outputs, while both trying to align with an extreme feature (e.g., the round line which is the overly abstract representation of the cat body in the first case of Figure \ref{fig:ablation}), the 'Dir' model manifests the feature as a curvy-shape background artifact while the 'NCE+aug' variant creates an exaggerated feature within object. This further underlines the regularizing effect of the CLIP-NCE energy. We thus conclude that using CLIP-NCE energy is preferable when applying our method to real-life sketches.

\textbf{Real-life sketches. }
We test our method with real-life sketches using only one reference sketch each case and show the results in Figure \ref{fig:handsketch}. We experiment on the same set of hand sketches taken from QuickDraw as GANSketch (first four in each category). Additionally, we pick sketches from the Sketchy dataset to verify our method on a wider range of sketch styles and object poses. We report the success cases in Figure \ref{fig:handsketch}. We first note that our method is able to generate realistic and sketch-conforming images in cases where GANSketch reports to fail even with multiple sketch inputs (e.g., row 3). Moreover, the output images shows a clear tendency of aligning high-level semantic concepts (e.g. pose and position). We further provide a qualitative comparison against baseline hand sketch results in Figure \ref{fig:compare}. The GANSketch images are generated from their published models and using the same set of input noises with truncation $\Phi =0.5$. We observe that our model outputs are of better qualities and exhibits semantic consistency in the object-level content. 

\section{Applications}
Apart from multi-modal sketch-guided synthesis, our model also could fit into the general scope of image editing. Since our method learns a conditional distribution without modifying the source GAN weights, latent-editing methods naturally remains compatible. We demonstrate the outcomes of applying latent-editing to our controlled models. We then show that to edit a natural image to resemble the sketch can be achieved through inverting it as the style latent and injecting it into a randomly sampled content latent from our learned distribution.   

\textbf{Latent-space editing. } One outstanding merit of using StyleGAN for graphic applications is its disentangled latent space, which enables intended editing to be conducted simply via perturbations along pre-identified directions in one of its intermediate latent space ($W$, $W+$ or $S$). As we do not modify the source model and consequently all of its intermediate latent space remaining unchanged, any latent-editing directions identified within the source model still apply, obviating the need to run direction discovery algorithms again. We verify this by applying  editing directions reported in \cite{harkonen2020ganspace} to random images sampled from our learned distribution and show that the edited result exhibit both sketch-controlled and latent-edited properties, as shown in \ref{fig:latentedit}.

\textbf{Natural image editing. } Our trained model controls the content of an output sample by a learned latent distribution conditioned on sketch and the style separately by an additional style latent drawn from the unchanged source model through style-mixing. Therefore it is possible to edit a natural image by first invert it as the style latent, then inject into a content latent randomly sampled from our learned distribution. To illustrate how this works, we first project real images to a $w'+$ latent using state-of-art inversion algorithm restyle \cite{restyle}. Same as the procedure in \ref{sec:stylemix}, we control the first 4 layers of the synthesis network with the content latent while using $w'+$ to control the rest; the result is shown in \ref{fig:realinvert}. We observed that the images are essentially edited versions of the original natural images with transformed pose and shape conforming to the initial sketch.

\section{Discussion}
In this work, we present a novel method for one shot sketch-conditioned image synthesis by leveraging pre-trained StyleGANs and CLIP. Our method do not required any image or sketch dataset, and the only user input is one hand-drawn sketch and a class-level neutral text description, leaving no extra hurdle for real-life application.

However, there are many room for further improvement. First, despite the capability to accommodate a wide range of sketch styles and entity poses as we have demonstrated, our method can not guarantee to work for all sketches. As observed in \cite{wu2022generative}, as the framework tackles conditional generation as mode-seeking in the source model’s data space, the generation tend to fail if the target mode is not covered by the pre-trained model. As shown in fig6, despite succeed in aligning with some semantic attributes from the sketch, our method fails to find a mode that matches the actual sketch concept entirely while maintaining photorealism. Second, our method relies on StyleGAN’s intrinsic structure. Finally, despite observing in the result a trade-off between realism and sketch-fitting, we have not provided any explicit way to control the degree of output realism, as \cite{cheng2023adaptively} did. We hope to leave this intriguing issue for future works.


\bibliographystyle{ACM-Reference-Format}
\bibliography{reference}


\begin{thebibliography}{44}


\ifx \showCODEN    \undefined \def \showCODEN     #1{\unskip}     \fi
\ifx \showDOI      \undefined \def \showDOI       #1{#1}\fi
\ifx \showISBNx    \undefined \def \showISBNx     #1{\unskip}     \fi
\ifx \showISBNxiii \undefined \def \showISBNxiii  #1{\unskip}     \fi
\ifx \showISSN     \undefined \def \showISSN      #1{\unskip}     \fi
\ifx \showLCCN     \undefined \def \showLCCN      #1{\unskip}     \fi
\ifx \shownote     \undefined \def \shownote      #1{#1}          \fi
\ifx \showarticletitle \undefined \def \showarticletitle #1{#1}   \fi
\ifx \showURL      \undefined \def \showURL       {\relax}        \fi
\providecommand\bibfield[2]{#2}
\providecommand\bibinfo[2]{#2}
\providecommand\natexlab[1]{#1}
\providecommand\showeprint[2][]{arXiv:#2}

\bibitem[Abdal et~al\mbox{.}(2021)]%
        {abdal2021styleflow}
\bibfield{author}{\bibinfo{person}{Rameen Abdal}, \bibinfo{person}{Peihao Zhu}, \bibinfo{person}{Niloy~J Mitra}, {and} \bibinfo{person}{Peter Wonka}.} \bibinfo{year}{2021}\natexlab{}.
\newblock \showarticletitle{Styleflow: Attribute-conditioned exploration of stylegan-generated images using conditional continuous normalizing flows}.
\newblock \bibinfo{journal}{\emph{ACM Transactions on Graphics (ToG)}} \bibinfo{volume}{40}, \bibinfo{number}{3} (\bibinfo{year}{2021}), \bibinfo{pages}{1--21}.
\newblock


\bibitem[Alaluf et~al\mbox{.}(2021)]%
        {restyle}
\bibfield{author}{\bibinfo{person}{Yuval Alaluf}, \bibinfo{person}{Or Patashnik}, {and} \bibinfo{person}{Daniel Cohen-Or}.} \bibinfo{year}{2021}\natexlab{}.
\newblock \showarticletitle{Restyle: A residual-based stylegan encoder via iterative refinement}. In \bibinfo{booktitle}{\emph{Proceedings of the IEEE/CVF International Conference on Computer Vision}}. \bibinfo{pages}{6711--6720}.
\newblock


\bibitem[Brooks et~al\mbox{.}(2023)]%
        {Brooks_2023_CVPR}
\bibfield{author}{\bibinfo{person}{Tim Brooks}, \bibinfo{person}{Aleksander Holynski}, {and} \bibinfo{person}{Alexei~A. Efros}.} \bibinfo{year}{2023}\natexlab{}.
\newblock \showarticletitle{InstructPix2Pix: Learning To Follow Image Editing Instructions}. In \bibinfo{booktitle}{\emph{Proceedings of the IEEE/CVF Conference on Computer Vision and Pattern Recognition (CVPR)}}. \bibinfo{pages}{18392--18402}.
\newblock


\bibitem[Chan et~al\mbox{.}(2022)]%
        {chan2022learning}
\bibfield{author}{\bibinfo{person}{Caroline Chan}, \bibinfo{person}{Fr{\'e}do Durand}, {and} \bibinfo{person}{Phillip Isola}.} \bibinfo{year}{2022}\natexlab{}.
\newblock \showarticletitle{Learning to generate line drawings that convey geometry and semantics}. In \bibinfo{booktitle}{\emph{Proceedings of the IEEE/CVF Conference on Computer Vision and Pattern Recognition}}. \bibinfo{pages}{7915--7925}.
\newblock


\bibitem[Cheema et~al\mbox{.}(2012)]%
        {cheema2012quickdraw}
\bibfield{author}{\bibinfo{person}{Salman Cheema}, \bibinfo{person}{Sumit Gulwani}, {and} \bibinfo{person}{Joseph LaViola}.} \bibinfo{year}{2012}\natexlab{}.
\newblock \showarticletitle{QuickDraw: improving drawing experience for geometric diagrams}. In \bibinfo{booktitle}{\emph{Proceedings of the SIGCHI Conference on Human Factors in Computing Systems}}. \bibinfo{pages}{1037--1064}.
\newblock


\bibitem[Chen and Hays(2018)]%
        {chen2018sketchygan}
\bibfield{author}{\bibinfo{person}{Wengling Chen} {and} \bibinfo{person}{James Hays}.} \bibinfo{year}{2018}\natexlab{}.
\newblock \showarticletitle{Sketchygan: Towards diverse and realistic sketch to image synthesis}. In \bibinfo{booktitle}{\emph{Proceedings of the IEEE Conference on Computer Vision and Pattern Recognition}}. \bibinfo{pages}{9416--9425}.
\newblock


\bibitem[Cheng et~al\mbox{.}(2023)]%
        {cheng2023adaptively}
\bibfield{author}{\bibinfo{person}{Shin-I Cheng}, \bibinfo{person}{Yu-Jie Chen}, \bibinfo{person}{Wei-Chen Chiu}, \bibinfo{person}{Hung-Yu Tseng}, {and} \bibinfo{person}{Hsin-Ying Lee}.} \bibinfo{year}{2023}\natexlab{}.
\newblock \showarticletitle{Adaptively-realistic image generation from stroke and sketch with diffusion model}. In \bibinfo{booktitle}{\emph{Proceedings of the IEEE/CVF Winter Conference on Applications of Computer Vision}}. \bibinfo{pages}{4054--4062}.
\newblock


\bibitem[Gal et~al\mbox{.}(2022)]%
        {gal2022stylegan}
\bibfield{author}{\bibinfo{person}{Rinon Gal}, \bibinfo{person}{Or Patashnik}, \bibinfo{person}{Haggai Maron}, \bibinfo{person}{Amit~H Bermano}, \bibinfo{person}{Gal Chechik}, {and} \bibinfo{person}{Daniel Cohen-Or}.} \bibinfo{year}{2022}\natexlab{}.
\newblock \showarticletitle{StyleGAN-NADA: CLIP-guided domain adaptation of image generators}.
\newblock \bibinfo{journal}{\emph{ACM Transactions on Graphics (TOG)}} \bibinfo{volume}{41}, \bibinfo{number}{4} (\bibinfo{year}{2022}), \bibinfo{pages}{1--13}.
\newblock


\bibitem[Goodfellow et~al\mbox{.}(2014)]%
        {goodfellow2014generative}
\bibfield{author}{\bibinfo{person}{Ian Goodfellow}, \bibinfo{person}{Jean Pouget-Abadie}, \bibinfo{person}{Mehdi Mirza}, \bibinfo{person}{Bing Xu}, \bibinfo{person}{David Warde-Farley}, \bibinfo{person}{Sherjil Ozair}, \bibinfo{person}{Aaron Courville}, {and} \bibinfo{person}{Yoshua Bengio}.} \bibinfo{year}{2014}\natexlab{}.
\newblock \showarticletitle{Generative adversarial nets}.
\newblock \bibinfo{journal}{\emph{Advances in neural information processing systems}}  \bibinfo{volume}{27} (\bibinfo{year}{2014}).
\newblock


\bibitem[H{\"a}rk{\"o}nen et~al\mbox{.}(2020)]%
        {harkonen2020ganspace}
\bibfield{author}{\bibinfo{person}{Erik H{\"a}rk{\"o}nen}, \bibinfo{person}{Aaron Hertzmann}, \bibinfo{person}{Jaakko Lehtinen}, {and} \bibinfo{person}{Sylvain Paris}.} \bibinfo{year}{2020}\natexlab{}.
\newblock \showarticletitle{Ganspace: Discovering interpretable gan controls}.
\newblock \bibinfo{journal}{\emph{Advances in neural information processing systems}}  \bibinfo{volume}{33} (\bibinfo{year}{2020}), \bibinfo{pages}{9841--9850}.
\newblock


\bibitem[Heusel et~al\mbox{.}(2017)]%
        {fid}
\bibfield{author}{\bibinfo{person}{Martin Heusel}, \bibinfo{person}{Hubert Ramsauer}, \bibinfo{person}{Thomas Unterthiner}, \bibinfo{person}{Bernhard Nessler}, {and} \bibinfo{person}{Sepp Hochreiter}.} \bibinfo{year}{2017}\natexlab{}.
\newblock \showarticletitle{Gans trained by a two time-scale update rule converge to a local nash equilibrium}.
\newblock \bibinfo{journal}{\emph{Advances in neural information processing systems}}  \bibinfo{volume}{30} (\bibinfo{year}{2017}).
\newblock


\bibitem[Karras et~al\mbox{.}(2019)]%
        {karras2019style}
\bibfield{author}{\bibinfo{person}{Tero Karras}, \bibinfo{person}{Samuli Laine}, {and} \bibinfo{person}{Timo Aila}.} \bibinfo{year}{2019}\natexlab{}.
\newblock \showarticletitle{A style-based generator architecture for generative adversarial networks}. In \bibinfo{booktitle}{\emph{Proceedings of the IEEE/CVF conference on computer vision and pattern recognition}}. \bibinfo{pages}{4401--4410}.
\newblock


\bibitem[Karras et~al\mbox{.}(2020)]%
        {karras2020analyzing}
\bibfield{author}{\bibinfo{person}{Tero Karras}, \bibinfo{person}{Samuli Laine}, \bibinfo{person}{Miika Aittala}, \bibinfo{person}{Janne Hellsten}, \bibinfo{person}{Jaakko Lehtinen}, {and} \bibinfo{person}{Timo Aila}.} \bibinfo{year}{2020}\natexlab{}.
\newblock \showarticletitle{Analyzing and improving the image quality of stylegan}. In \bibinfo{booktitle}{\emph{Proceedings of the IEEE/CVF conference on computer vision and pattern recognition}}. \bibinfo{pages}{8110--8119}.
\newblock


\bibitem[Koley et~al\mbox{.}(2023)]%
        {koley2023picture}
\bibfield{author}{\bibinfo{person}{Subhadeep Koley}, \bibinfo{person}{Ayan~Kumar Bhunia}, \bibinfo{person}{Aneeshan Sain}, \bibinfo{person}{Pinaki~Nath Chowdhury}, \bibinfo{person}{Tao Xiang}, {and} \bibinfo{person}{Yi-Zhe Song}.} \bibinfo{year}{2023}\natexlab{}.
\newblock \showarticletitle{Picture that Sketch: Photorealistic Image Generation from Abstract Sketches}. In \bibinfo{booktitle}{\emph{Proceedings of the IEEE/CVF Conference on Computer Vision and Pattern Recognition}}. \bibinfo{pages}{6850--6861}.
\newblock


\bibitem[Kwon and Ye(2023)]%
        {kwon2023one}
\bibfield{author}{\bibinfo{person}{Gihyun Kwon} {and} \bibinfo{person}{Jong~Chul Ye}.} \bibinfo{year}{2023}\natexlab{}.
\newblock \showarticletitle{One-shot adaptation of gan in just one clip}.
\newblock \bibinfo{journal}{\emph{IEEE Transactions on Pattern Analysis and Machine Intelligence}} (\bibinfo{year}{2023}).
\newblock


\bibitem[Kynk{\"a}{\"a}nniemi et~al\mbox{.}(2019)]%
        {pp}
\bibfield{author}{\bibinfo{person}{Tuomas Kynk{\"a}{\"a}nniemi}, \bibinfo{person}{Tero Karras}, \bibinfo{person}{Samuli Laine}, \bibinfo{person}{Jaakko Lehtinen}, {and} \bibinfo{person}{Timo Aila}.} \bibinfo{year}{2019}\natexlab{}.
\newblock \showarticletitle{Improved precision and recall metric for assessing generative models}.
\newblock \bibinfo{journal}{\emph{Advances in Neural Information Processing Systems}}  \bibinfo{volume}{32} (\bibinfo{year}{2019}).
\newblock


\bibitem[LeCun et~al\mbox{.}(2006)]%
        {lecun2006tutorial}
\bibfield{author}{\bibinfo{person}{Yann LeCun}, \bibinfo{person}{Sumit Chopra}, \bibinfo{person}{Raia Hadsell}, \bibinfo{person}{M Ranzato}, {and} \bibinfo{person}{Fujie Huang}.} \bibinfo{year}{2006}\natexlab{}.
\newblock \showarticletitle{A tutorial on energy-based learning}.
\newblock \bibinfo{journal}{\emph{Predicting structured data}} \bibinfo{volume}{1}, \bibinfo{number}{0} (\bibinfo{year}{2006}).
\newblock


\bibitem[Lee et~al\mbox{.}(2023)]%
        {Lee_2023_CVPR}
\bibfield{author}{\bibinfo{person}{Hyundo Lee}, \bibinfo{person}{Inwoo Hwang}, \bibinfo{person}{Hyunsung Go}, \bibinfo{person}{Won-Seok Choi}, \bibinfo{person}{Kibeom Kim}, {and} \bibinfo{person}{Byoung-Tak Zhang}.} \bibinfo{year}{2023}\natexlab{}.
\newblock \showarticletitle{Learning Geometry-Aware Representations by Sketching}. In \bibinfo{booktitle}{\emph{Proceedings of the IEEE/CVF Conference on Computer Vision and Pattern Recognition (CVPR)}}. \bibinfo{pages}{23315--23326}.
\newblock


\bibitem[Li et~al\mbox{.}(2019b)]%
        {li2019controllable}
\bibfield{author}{\bibinfo{person}{Bowen Li}, \bibinfo{person}{Xiaojuan Qi}, \bibinfo{person}{Thomas Lukasiewicz}, {and} \bibinfo{person}{Philip Torr}.} \bibinfo{year}{2019}\natexlab{b}.
\newblock \showarticletitle{Controllable text-to-image generation}.
\newblock \bibinfo{journal}{\emph{Advances in Neural Information Processing Systems}}  \bibinfo{volume}{32} (\bibinfo{year}{2019}).
\newblock


\bibitem[Li et~al\mbox{.}(2019a)]%
        {li2019photo}
\bibfield{author}{\bibinfo{person}{Mengtian Li}, \bibinfo{person}{Zhe Lin}, \bibinfo{person}{Radomir Mech}, \bibinfo{person}{Ersin Yumer}, {and} \bibinfo{person}{Deva Ramanan}.} \bibinfo{year}{2019}\natexlab{a}.
\newblock \showarticletitle{Photo-sketching: Inferring contour drawings from images}. In \bibinfo{booktitle}{\emph{2019 IEEE Winter Conference on Applications of Computer Vision (WACV)}}. IEEE, \bibinfo{pages}{1403--1412}.
\newblock


\bibitem[Liu et~al\mbox{.}(2021)]%
        {liu2021self}
\bibfield{author}{\bibinfo{person}{Bingchen Liu}, \bibinfo{person}{Yizhe Zhu}, \bibinfo{person}{Kunpeng Song}, {and} \bibinfo{person}{Ahmed Elgammal}.} \bibinfo{year}{2021}\natexlab{}.
\newblock \showarticletitle{Self-supervised sketch-to-image synthesis}. In \bibinfo{booktitle}{\emph{Proceedings of the AAAI conference on artificial intelligence}}, Vol.~\bibinfo{volume}{35}. \bibinfo{pages}{2073--2081}.
\newblock


\bibitem[Liu et~al\mbox{.}(2020)]%
        {liu2020unsupervised}
\bibfield{author}{\bibinfo{person}{Runtao Liu}, \bibinfo{person}{Qian Yu}, {and} \bibinfo{person}{Stella~X Yu}.} \bibinfo{year}{2020}\natexlab{}.
\newblock \showarticletitle{Unsupervised sketch to photo synthesis}. In \bibinfo{booktitle}{\emph{Computer Vision--ECCV 2020: 16th European Conference, Glasgow, UK, August 23--28, 2020, Proceedings, Part III 16}}. Springer, \bibinfo{pages}{36--52}.
\newblock


\bibitem[Nguyen et~al\mbox{.}(2017)]%
        {plug}
\bibfield{author}{\bibinfo{person}{Anh Nguyen}, \bibinfo{person}{Jeff Clune}, \bibinfo{person}{Yoshua Bengio}, \bibinfo{person}{Alexey Dosovitskiy}, {and} \bibinfo{person}{Jason Yosinski}.} \bibinfo{year}{2017}\natexlab{}.
\newblock \showarticletitle{Plug \& play generative networks: Conditional iterative generation of images in latent space}. In \bibinfo{booktitle}{\emph{Proceedings of the IEEE conference on computer vision and pattern recognition}}. \bibinfo{pages}{4467--4477}.
\newblock


\bibitem[Nichol et~al\mbox{.}(2021)]%
        {nichol2021glide}
\bibfield{author}{\bibinfo{person}{Alex Nichol}, \bibinfo{person}{Prafulla Dhariwal}, \bibinfo{person}{Aditya Ramesh}, \bibinfo{person}{Pranav Shyam}, \bibinfo{person}{Pamela Mishkin}, \bibinfo{person}{Bob McGrew}, \bibinfo{person}{Ilya Sutskever}, {and} \bibinfo{person}{Mark Chen}.} \bibinfo{year}{2021}\natexlab{}.
\newblock \showarticletitle{Glide: Towards photorealistic image generation and editing with text-guided diffusion models}.
\newblock \bibinfo{journal}{\emph{arXiv preprint arXiv:2112.10741}} (\bibinfo{year}{2021}).
\newblock


\bibitem[Nie et~al\mbox{.}(2021)]%
        {nie2021controllable}
\bibfield{author}{\bibinfo{person}{Weili Nie}, \bibinfo{person}{Arash Vahdat}, {and} \bibinfo{person}{Anima Anandkumar}.} \bibinfo{year}{2021}\natexlab{}.
\newblock \showarticletitle{Controllable and compositional generation with latent-space energy-based models}.
\newblock \bibinfo{journal}{\emph{Advances in Neural Information Processing Systems}}  \bibinfo{volume}{34} (\bibinfo{year}{2021}), \bibinfo{pages}{13497--13510}.
\newblock


\bibitem[No{\'e} et~al\mbox{.}(2019)]%
        {inn}
\bibfield{author}{\bibinfo{person}{Frank No{\'e}}, \bibinfo{person}{Simon Olsson}, \bibinfo{person}{Jonas K{\"o}hler}, {and} \bibinfo{person}{Hao Wu}.} \bibinfo{year}{2019}\natexlab{}.
\newblock \showarticletitle{Boltzmann generators: Sampling equilibrium states of many-body systems with deep learning}.
\newblock \bibinfo{journal}{\emph{Science}} \bibinfo{volume}{365}, \bibinfo{number}{6457} (\bibinfo{year}{2019}), \bibinfo{pages}{eaaw1147}.
\newblock


\bibitem[Oord et~al\mbox{.}(2018)]%
        {infonce}
\bibfield{author}{\bibinfo{person}{Aaron van~den Oord}, \bibinfo{person}{Yazhe Li}, {and} \bibinfo{person}{Oriol Vinyals}.} \bibinfo{year}{2018}\natexlab{}.
\newblock \showarticletitle{Representation learning with contrastive predictive coding}.
\newblock \bibinfo{journal}{\emph{arXiv preprint arXiv:1807.03748}} (\bibinfo{year}{2018}).
\newblock


\bibitem[Parmar et~al\mbox{.}(2022)]%
        {parmar2021cleanfid}
\bibfield{author}{\bibinfo{person}{Gaurav Parmar}, \bibinfo{person}{Richard Zhang}, {and} \bibinfo{person}{Jun-Yan Zhu}.} \bibinfo{year}{2022}\natexlab{}.
\newblock \showarticletitle{On Aliased Resizing and Surprising Subtleties in GAN Evaluation}. In \bibinfo{booktitle}{\emph{CVPR}}.
\newblock


\bibitem[Patashnik et~al\mbox{.}(2021)]%
        {patashnik2021styleclip}
\bibfield{author}{\bibinfo{person}{Or Patashnik}, \bibinfo{person}{Zongze Wu}, \bibinfo{person}{Eli Shechtman}, \bibinfo{person}{Daniel Cohen-Or}, {and} \bibinfo{person}{Dani Lischinski}.} \bibinfo{year}{2021}\natexlab{}.
\newblock \showarticletitle{Styleclip: Text-driven manipulation of stylegan imagery}. In \bibinfo{booktitle}{\emph{Proceedings of the IEEE/CVF International Conference on Computer Vision}}. \bibinfo{pages}{2085--2094}.
\newblock


\bibitem[Radford et~al\mbox{.}(2021)]%
        {clip}
\bibfield{author}{\bibinfo{person}{Alec Radford}, \bibinfo{person}{Jong~Wook Kim}, \bibinfo{person}{Chris Hallacy}, \bibinfo{person}{Aditya Ramesh}, \bibinfo{person}{Gabriel Goh}, \bibinfo{person}{Sandhini Agarwal}, \bibinfo{person}{Girish Sastry}, \bibinfo{person}{Amanda Askell}, \bibinfo{person}{Pamela Mishkin}, \bibinfo{person}{Jack Clark}, {et~al\mbox{.}}} \bibinfo{year}{2021}\natexlab{}.
\newblock \showarticletitle{Learning transferable visual models from natural language supervision}. In \bibinfo{booktitle}{\emph{International conference on machine learning}}. PMLR, \bibinfo{pages}{8748--8763}.
\newblock


\bibitem[Richardson et~al\mbox{.}(2021)]%
        {richardson2021encoding}
\bibfield{author}{\bibinfo{person}{Elad Richardson}, \bibinfo{person}{Yuval Alaluf}, \bibinfo{person}{Or Patashnik}, \bibinfo{person}{Yotam Nitzan}, \bibinfo{person}{Yaniv Azar}, \bibinfo{person}{Stav Shapiro}, {and} \bibinfo{person}{Daniel Cohen-Or}.} \bibinfo{year}{2021}\natexlab{}.
\newblock \showarticletitle{Encoding in style: a stylegan encoder for image-to-image translation}. In \bibinfo{booktitle}{\emph{Proceedings of the IEEE/CVF conference on computer vision and pattern recognition}}. \bibinfo{pages}{2287--2296}.
\newblock


\bibitem[Sangkloy et~al\mbox{.}(2016)]%
        {sangkloy2016sketchy}
\bibfield{author}{\bibinfo{person}{Patsorn Sangkloy}, \bibinfo{person}{Nathan Burnell}, \bibinfo{person}{Cusuh Ham}, {and} \bibinfo{person}{James Hays}.} \bibinfo{year}{2016}\natexlab{}.
\newblock \showarticletitle{The sketchy database: learning to retrieve badly drawn bunnies}.
\newblock \bibinfo{journal}{\emph{ACM Transactions on Graphics (TOG)}} \bibinfo{volume}{35}, \bibinfo{number}{4} (\bibinfo{year}{2016}), \bibinfo{pages}{1--12}.
\newblock


\bibitem[Sangkloy et~al\mbox{.}(2017)]%
        {sangkloy2017scribbler}
\bibfield{author}{\bibinfo{person}{Patsorn Sangkloy}, \bibinfo{person}{Jingwan Lu}, \bibinfo{person}{Chen Fang}, \bibinfo{person}{Fisher Yu}, {and} \bibinfo{person}{James Hays}.} \bibinfo{year}{2017}\natexlab{}.
\newblock \showarticletitle{Scribbler: Controlling deep image synthesis with sketch and color}. In \bibinfo{booktitle}{\emph{Proceedings of the IEEE conference on computer vision and pattern recognition}}. \bibinfo{pages}{5400--5409}.
\newblock


\bibitem[Shen et~al\mbox{.}(2020)]%
        {shen2020interpreting}
\bibfield{author}{\bibinfo{person}{Yujun Shen}, \bibinfo{person}{Jinjin Gu}, \bibinfo{person}{Xiaoou Tang}, {and} \bibinfo{person}{Bolei Zhou}.} \bibinfo{year}{2020}\natexlab{}.
\newblock \showarticletitle{Interpreting the latent space of gans for semantic face editing}. In \bibinfo{booktitle}{\emph{Proceedings of the IEEE/CVF conference on computer vision and pattern recognition}}. \bibinfo{pages}{9243--9252}.
\newblock


\bibitem[Shoshan et~al\mbox{.}(2021)]%
        {shoshan2021gan}
\bibfield{author}{\bibinfo{person}{Alon Shoshan}, \bibinfo{person}{Nadav Bhonker}, \bibinfo{person}{Igor Kviatkovsky}, {and} \bibinfo{person}{Gerard Medioni}.} \bibinfo{year}{2021}\natexlab{}.
\newblock \showarticletitle{Gan-control: Explicitly controllable gans}. In \bibinfo{booktitle}{\emph{Proceedings of the IEEE/CVF international conference on computer vision}}. \bibinfo{pages}{14083--14093}.
\newblock


\bibitem[Simonyan and Zisserman(2014)]%
        {vgg}
\bibfield{author}{\bibinfo{person}{Karen Simonyan} {and} \bibinfo{person}{Andrew Zisserman}.} \bibinfo{year}{2014}\natexlab{}.
\newblock \showarticletitle{Very deep convolutional networks for large-scale image recognition}.
\newblock \bibinfo{journal}{\emph{arXiv preprint arXiv:1409.1556}} (\bibinfo{year}{2014}).
\newblock


\bibitem[Wang et~al\mbox{.}(2021)]%
        {syog}
\bibfield{author}{\bibinfo{person}{Sheng-Yu Wang}, \bibinfo{person}{David Bau}, {and} \bibinfo{person}{Jun-Yan Zhu}.} \bibinfo{year}{2021}\natexlab{}.
\newblock \showarticletitle{Sketch your own gan}. In \bibinfo{booktitle}{\emph{Proceedings of the IEEE/CVF International Conference on Computer Vision}}. \bibinfo{pages}{14050--14060}.
\newblock


\bibitem[Wang et~al\mbox{.}(2018)]%
        {wang2018transferring}
\bibfield{author}{\bibinfo{person}{Yaxing Wang}, \bibinfo{person}{Chenshen Wu}, \bibinfo{person}{Luis Herranz}, \bibinfo{person}{Joost Van~de Weijer}, \bibinfo{person}{Abel Gonzalez-Garcia}, {and} \bibinfo{person}{Bogdan Raducanu}.} \bibinfo{year}{2018}\natexlab{}.
\newblock \showarticletitle{Transferring gans: generating images from limited data}. In \bibinfo{booktitle}{\emph{Proceedings of the European Conference on Computer Vision (ECCV)}}. \bibinfo{pages}{218--234}.
\newblock


\bibitem[Wu et~al\mbox{.}(2022)]%
        {wu2022generative}
\bibfield{author}{\bibinfo{person}{Chen~Henry Wu}, \bibinfo{person}{Saman Motamed}, \bibinfo{person}{Shaunak Srivastava}, {and} \bibinfo{person}{Fernando~D De~la Torre}.} \bibinfo{year}{2022}\natexlab{}.
\newblock \showarticletitle{Generative visual prompt: Unifying distributional control of pre-trained generative models}.
\newblock \bibinfo{journal}{\emph{Advances in Neural Information Processing Systems}}  \bibinfo{volume}{35} (\bibinfo{year}{2022}), \bibinfo{pages}{22422--22437}.
\newblock


\bibitem[Wu et~al\mbox{.}(2021)]%
        {wu2021stylespace}
\bibfield{author}{\bibinfo{person}{Zongze Wu}, \bibinfo{person}{Dani Lischinski}, {and} \bibinfo{person}{Eli Shechtman}.} \bibinfo{year}{2021}\natexlab{}.
\newblock \showarticletitle{Stylespace analysis: Disentangled controls for stylegan image generation}. In \bibinfo{booktitle}{\emph{Proceedings of the IEEE/CVF Conference on Computer Vision and Pattern Recognition}}. \bibinfo{pages}{12863--12872}.
\newblock


\bibitem[Yu et~al\mbox{.}(2015)]%
        {yu2015lsun}
\bibfield{author}{\bibinfo{person}{Fisher Yu}, \bibinfo{person}{Ari Seff}, \bibinfo{person}{Yinda Zhang}, \bibinfo{person}{Shuran Song}, \bibinfo{person}{Thomas Funkhouser}, {and} \bibinfo{person}{Jianxiong Xiao}.} \bibinfo{year}{2015}\natexlab{}.
\newblock \showarticletitle{Lsun: Construction of a large-scale image dataset using deep learning with humans in the loop}.
\newblock \bibinfo{journal}{\emph{arXiv preprint arXiv:1506.03365}} (\bibinfo{year}{2015}).
\newblock


\bibitem[Yu et~al\mbox{.}(2022)]%
        {cfclip}
\bibfield{author}{\bibinfo{person}{Yingchen Yu}, \bibinfo{person}{Fangneng Zhan}, \bibinfo{person}{Rongliang Wu}, \bibinfo{person}{Jiahui Zhang}, \bibinfo{person}{Shijian Lu}, \bibinfo{person}{Miaomiao Cui}, \bibinfo{person}{Xuansong Xie}, \bibinfo{person}{Xian-Sheng Hua}, {and} \bibinfo{person}{Chunyan Miao}.} \bibinfo{year}{2022}\natexlab{}.
\newblock \showarticletitle{Towards counterfactual image manipulation via clip}. In \bibinfo{booktitle}{\emph{Proceedings of the 30th ACM International Conference on Multimedia}}. \bibinfo{pages}{3637--3645}.
\newblock


\bibitem[Zhao et~al\mbox{.}(2020)]%
        {diffaug}
\bibfield{author}{\bibinfo{person}{Shengyu Zhao}, \bibinfo{person}{Zhijian Liu}, \bibinfo{person}{Ji Lin}, \bibinfo{person}{Jun-Yan Zhu}, {and} \bibinfo{person}{Song Han}.} \bibinfo{year}{2020}\natexlab{}.
\newblock \showarticletitle{Differentiable augmentation for data-efficient gan training}.
\newblock \bibinfo{journal}{\emph{Advances in neural information processing systems}}  \bibinfo{volume}{33} (\bibinfo{year}{2020}), \bibinfo{pages}{7559--7570}.
\newblock


\bibitem[Zhu et~al\mbox{.}(2021)]%
        {zhu2021mind}
\bibfield{author}{\bibinfo{person}{Peihao Zhu}, \bibinfo{person}{Rameen Abdal}, \bibinfo{person}{John Femiani}, {and} \bibinfo{person}{Peter Wonka}.} \bibinfo{year}{2021}\natexlab{}.
\newblock \showarticletitle{Mind the Gap: Domain Gap Control for Single Shot Domain Adaptation for Generative Adversarial Networks}. In \bibinfo{booktitle}{\emph{International Conference on Learning Representations}}.
\newblock


\end{thebibliography}
\clearpage
\appendix

\begin{table*}[!ht]
\begin{tabular}{@{}lllllllllll@{}}
\toprule
Model              & \multicolumn{2}{l}{Training Settings} & \multicolumn{2}{l}{standing cat}                                           & \multicolumn{2}{l}{horse rider}                                            & \multicolumn{2}{l}{horse on side}                                      & \multicolumn{2}{l}{garbled church}                                         \\ \midrule
                   & No. Samples           & Aug.          & Prec. $\uparrow$                                & Rec.$\uparrow$                                 & Prec.$\uparrow$                                & Rec.$\uparrow$                                 & Prec. $\uparrow$                               & Rec. $\uparrow$                                & Prec.$\uparrow$                                & Rec.$\uparrow$                                 \\ \midrule
Original           & N.A                   &               & {\color[HTML]{9B9B9B} \textit{0.21}} & {\color[HTML]{9B9B9B} \textit{0.54}} & {\color[HTML]{9B9B9B} \textit{0.33}} & {\color[HTML]{9B9B9B} \textit{0.57}} & {\color[HTML]{9B9B9B} \textit{0.22}} & {\color[HTML]{9B9B9B} \textit{0.63}} & {\color[HTML]{9B9B9B} \textit{0.46}} & {\color[HTML]{9B9B9B} \textit{0.49}} \\ \midrule
GanSketch(w aug)   & 30                    & \checkmark             & 0.50                                 & \textbf{0.41}                        & \textbf{0.44}                        & 0.39                                 & \textbf{0.50}                        & 0.50                                 & 0.46                                 & \textbf{0.51}                        \\
GanSketch(w/o aug) & 30                    &               & 0.65                                 & 0.20                                 & 0.42                                 & 0.49                                 & 0.42                                 & \textbf{0.52}                        & 0.48                                 & 0.48                                 \\ \midrule
Ours(w/aug)        & 1                     & \checkmark             & 0.62                                 & 0.25                                 & 0.26                                 & \textbf{0.49}                        & 0.41                                 & 0.45                                 & \textbf{0.64}                        & 0.16                                 \\
Ours(w/o aug)      & 1                     &               & \textbf{0.69}                        & 0.32                                 & 0.32                                 & 0.48                                 & 0.40                                 & 0.34                                 & 0.63                                 & 0.34                                 \\ \bottomrule
\end{tabular}
\caption{\textbf{Precision and Recall metrics on synthesized sketches. } We report the Precision and Recall metrics of our models against the original pre-trained StyleGANs and the baseline (GANSketch), tested with the synthesized sketches. We report both the augmented (w/aug) and unaugmented (w/o aug) variants of our method and baseline method. Our models are trained with single input sketch while baseline ones are trained with 30. The GANSketch scores are directly taken from their report. For both precision and recall scores, the higher the better. We mark that highest score in \textbf{black}.}
\label{tablepp}
\end{table*}

\section{Implementation Details}
\textbf{Training details. }
We train $f_{theta}$ using the same training hyperparameters as \cite{wu2022generative} except gradient accumulation, batch size, and number of epochs. We train each photosketch cases for 5 epochs and each hand-sketch case for 10 epochs. All models are trained with no gradient accumulation and batch size of one, which we find essential for our training.  We apply truncation $\Phi=0.5$ for all cases when sampling random style latent for style-mixing. The sampled content latent is not truncated during the training.

\textbf{Hyperparameters. }
For energy weight factor $\lambda$, we did a grid search in a search space of [1000,2000,5000], and find 2000 to generally work for all case except 'standing cat', which is trained with 5000. We also follow \cite{cfclip} to set the infoNCE temperature to 0.1. 

\textbf{Evaluation Procedure. }
We follow the same evaluation scheme as GANSketch \cite{syog}. For quantitative evaluation, we sample 2500 latents from our models without truncation. Each latent is injected with a randomly-sampled style latent before transformed to a 256x256 image. The evaluation set contains 2500 hand-pick images at resolution 256. The FID scores are calculated with the CleanFID library \cite{parmar2021cleanfid}. For qualitative cases, all examples are generated with truncation $\Phi=0.5$ and style-mixing.
\section{Additional Results. }
\textbf{Other evaluation metrics. }
We additionally report the Precision and Recall metrics \cite{pp}. The precision score indicates the proportion in generated samples that resemble the target real images in evaluation set when comparing VGG features \cite{vgg}, while the recall indicates the proportion of real data that resemble the generated samples. The decreased recall indicates a smaller coverage of original modes, which is as expected, since we aim to seek specific modes from the source data space. Our model exhibits a increased precision in the majority of the test cases, indicating its ability to refine a conditioned distribution that resembles the target distribution defined by the evaluation dataset. The only exception being the horse side case, as we observe that extreme object scale (e.g. too large or small) is challenging for our method.       

\begin{figure*}[h]
  \centering
  \includegraphics[width=\linewidth]{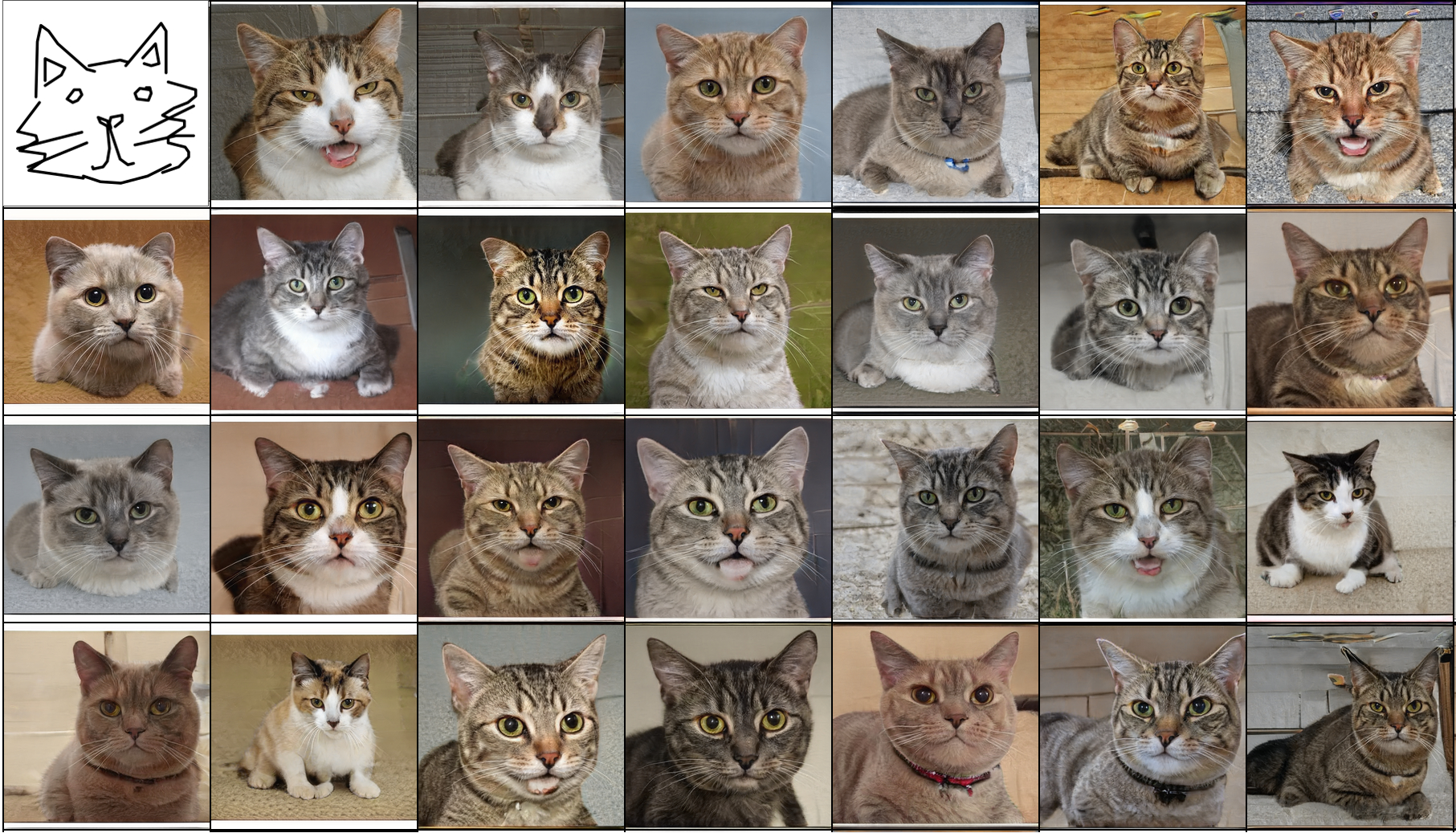}
  \caption{\textbf{Uncurated results }}
\label{fig:uncuratedcat0}
\end{figure*}

\begin{figure*}[h]
  \centering
  \includegraphics[width=\linewidth]{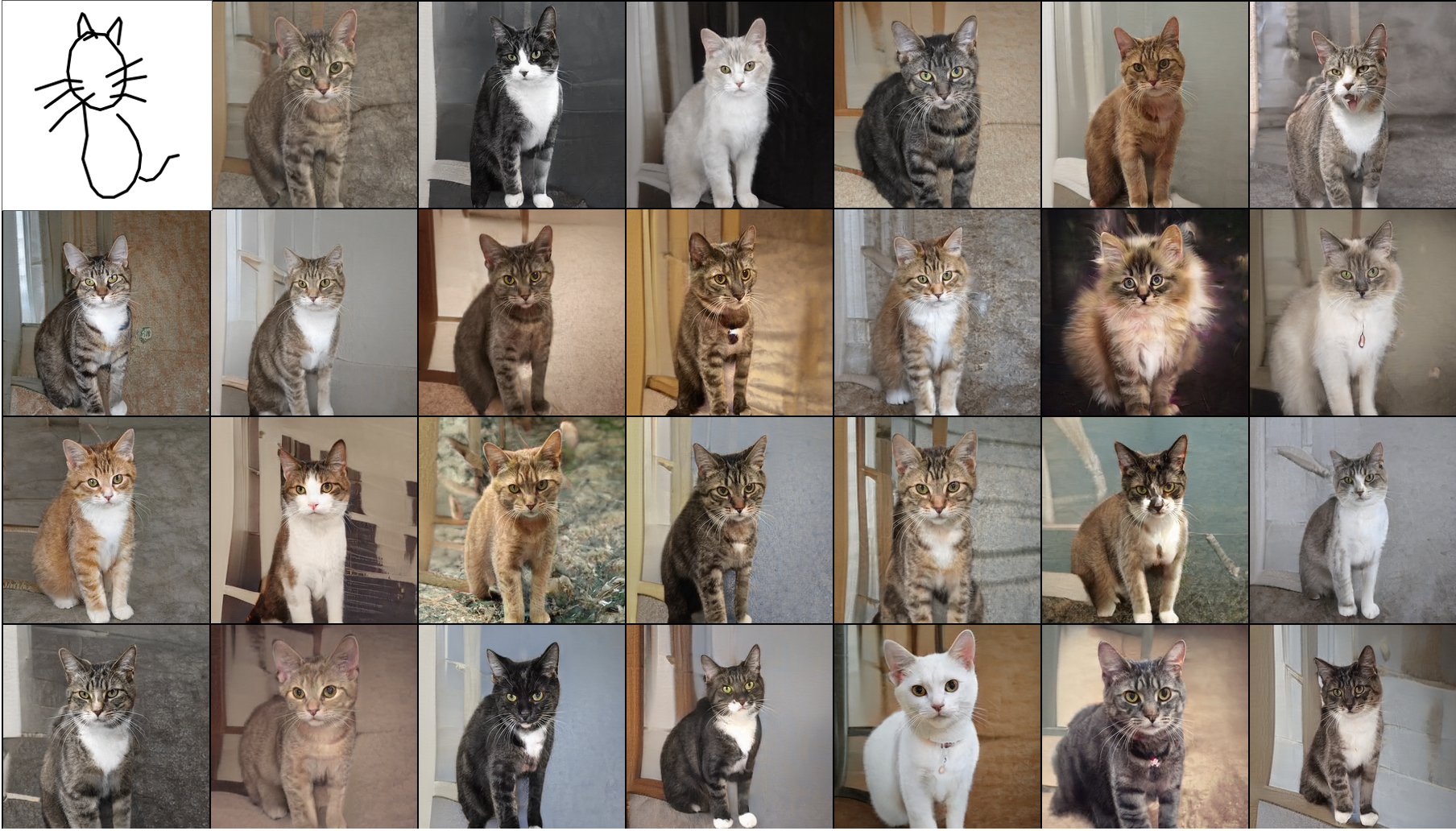}
  \caption{\textbf{Uncurated results }}
  \label{fig:uncuratedcat1}
\end{figure*}

\begin{figure*}[h]
  \centering
  \includegraphics[width=\linewidth]{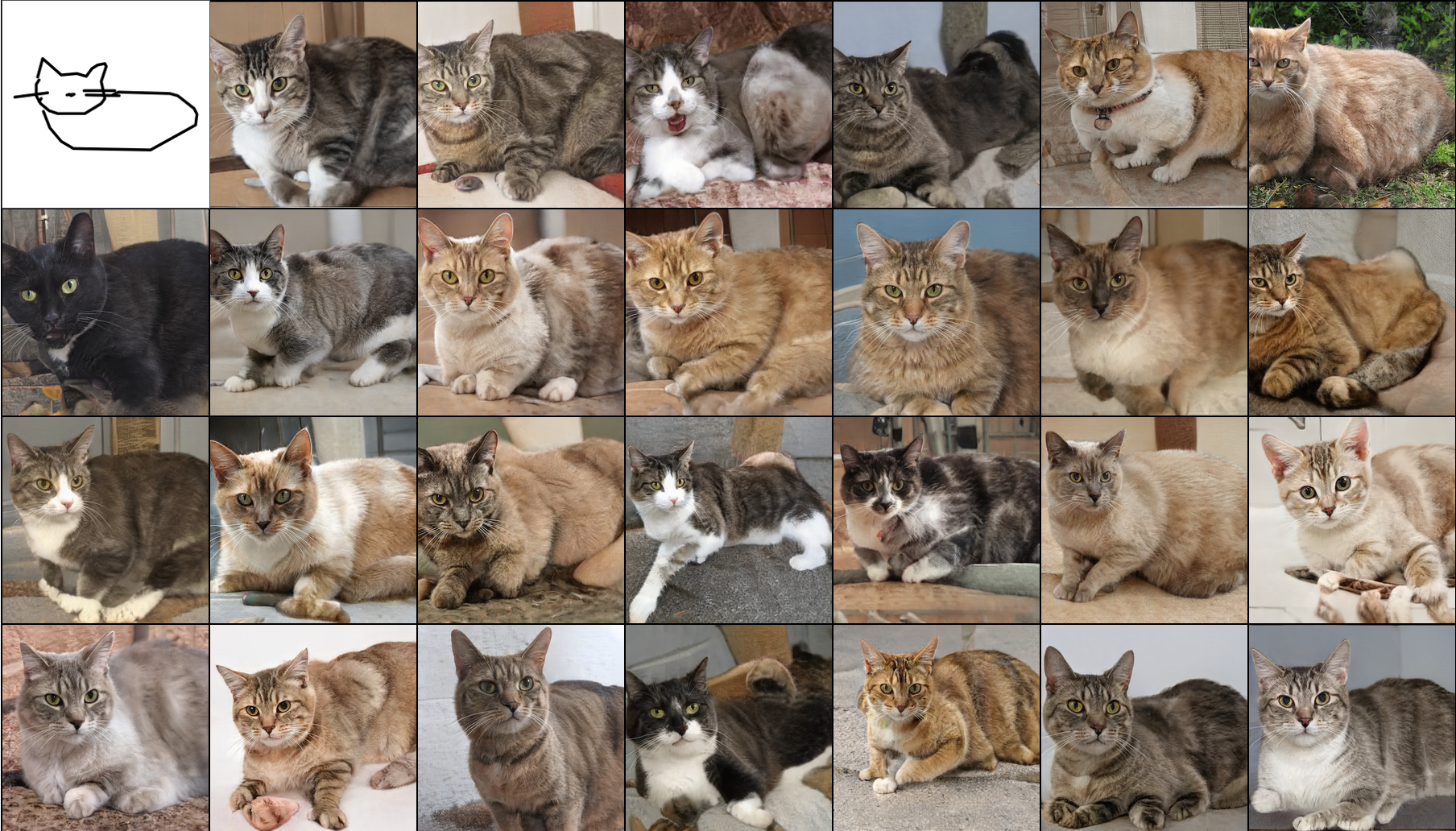}
  \caption{\textbf{Uncurated results }}
  \label{fig:uncuratedcat2}
\end{figure*}

\begin{figure*}[h]
  \centering
  \includegraphics[width=\linewidth]{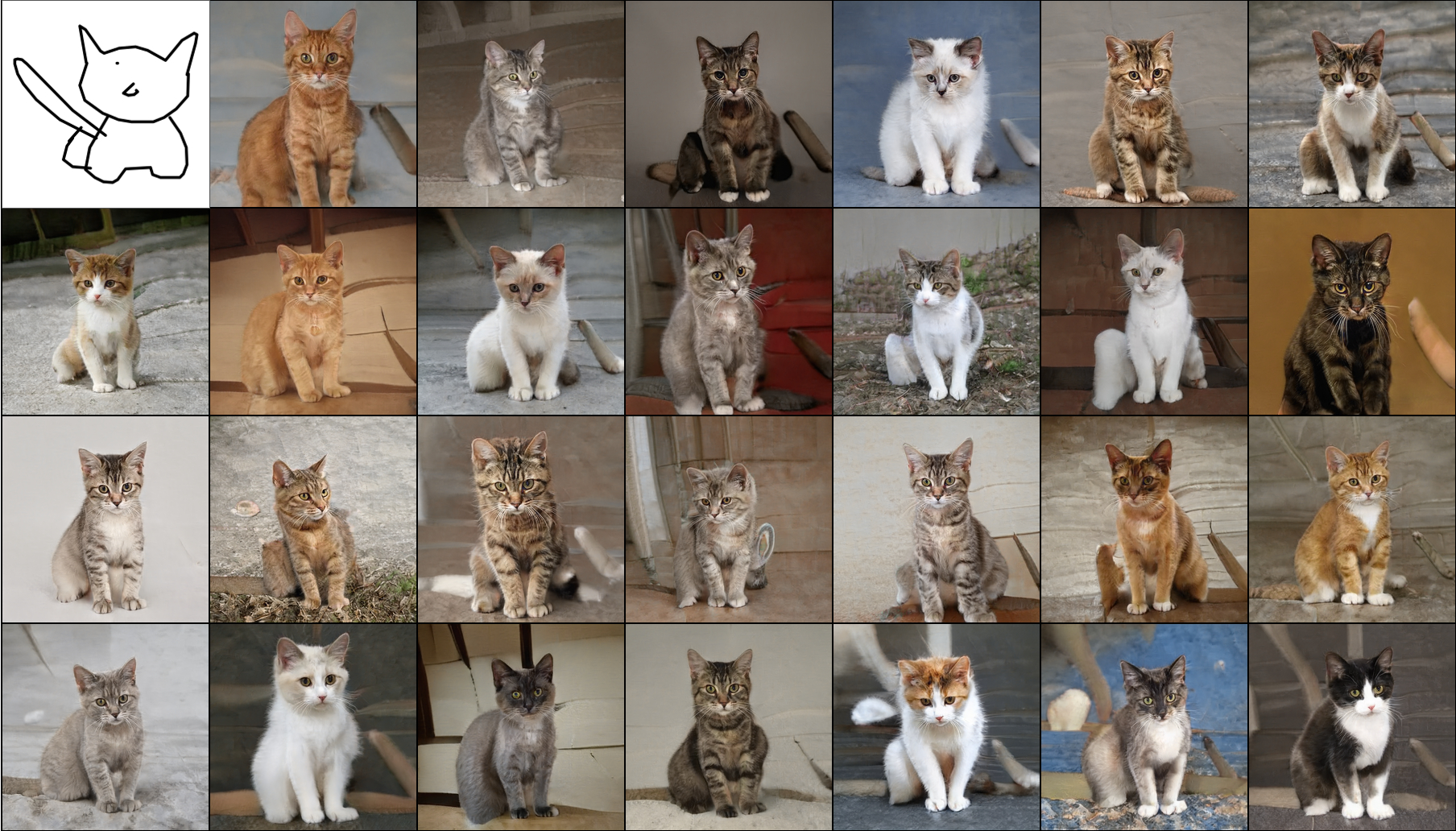}
  \caption{\textbf{Uncurated results }}
  \label{fig:uncuratedcat3}
\end{figure*}

\begin{figure*}[h]
  \centering
  \includegraphics[width=\linewidth]{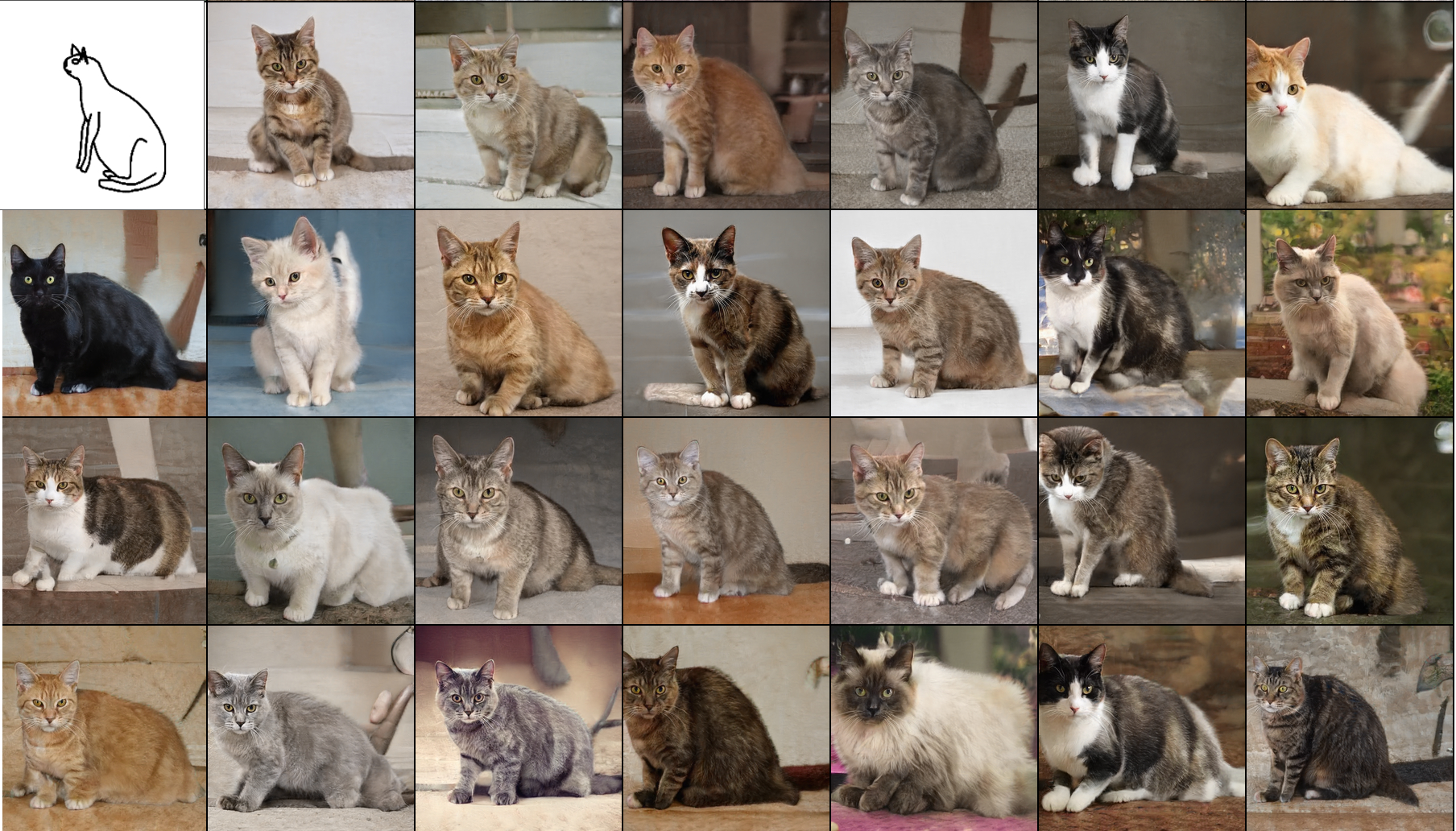}
  \caption{\textbf{Uncurated results }}
  \label{fig:uncuratedcatsketchy1}
\end{figure*}

\begin{figure*}[h]
  \centering
  \includegraphics[width=\linewidth]{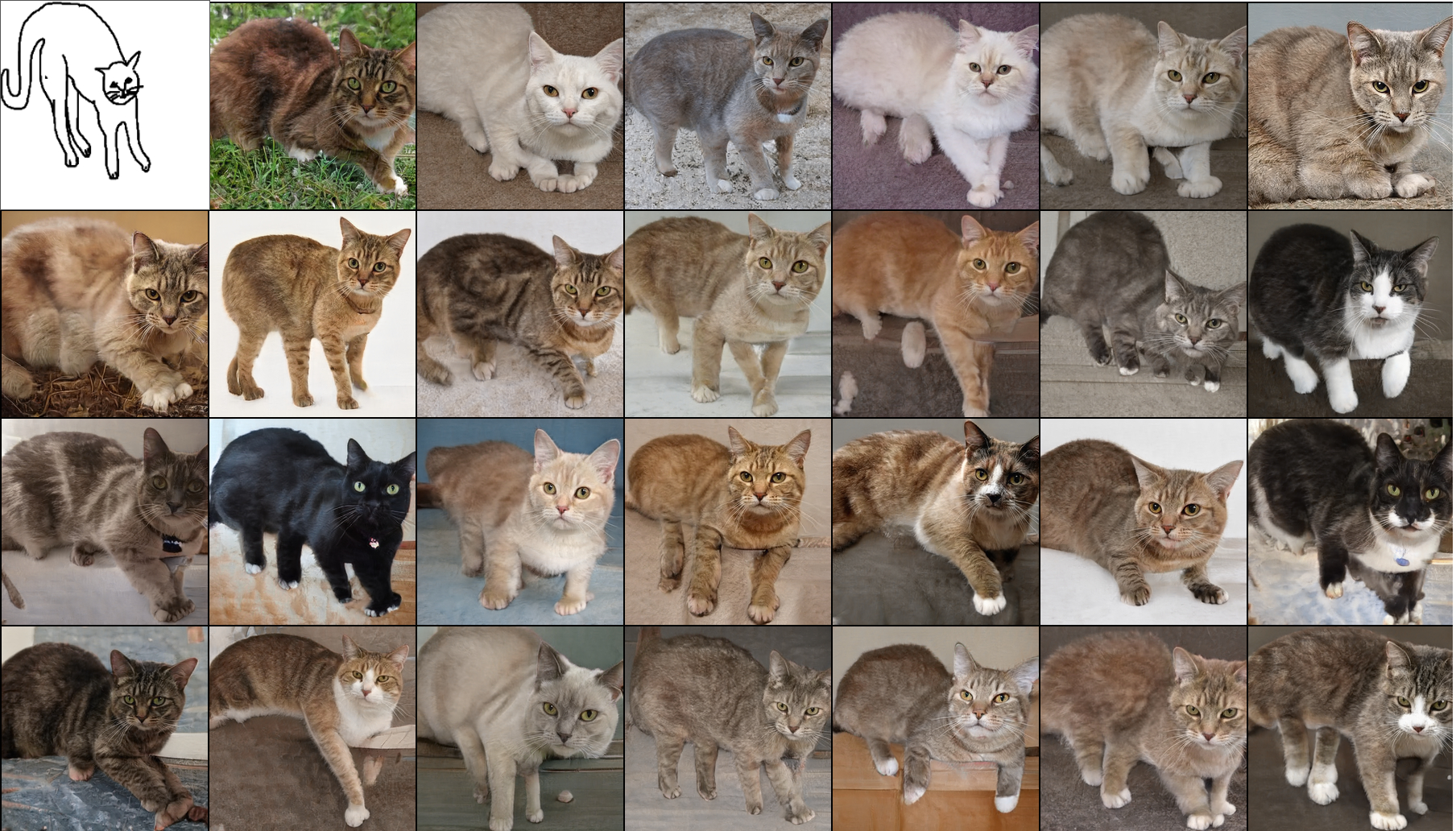}
  \caption{\textbf{Uncurated results }}
  \label{fig:uncuratedcatsketchy2}
\end{figure*}

\begin{figure*}[h]
  \centering
  \includegraphics[width=\linewidth]{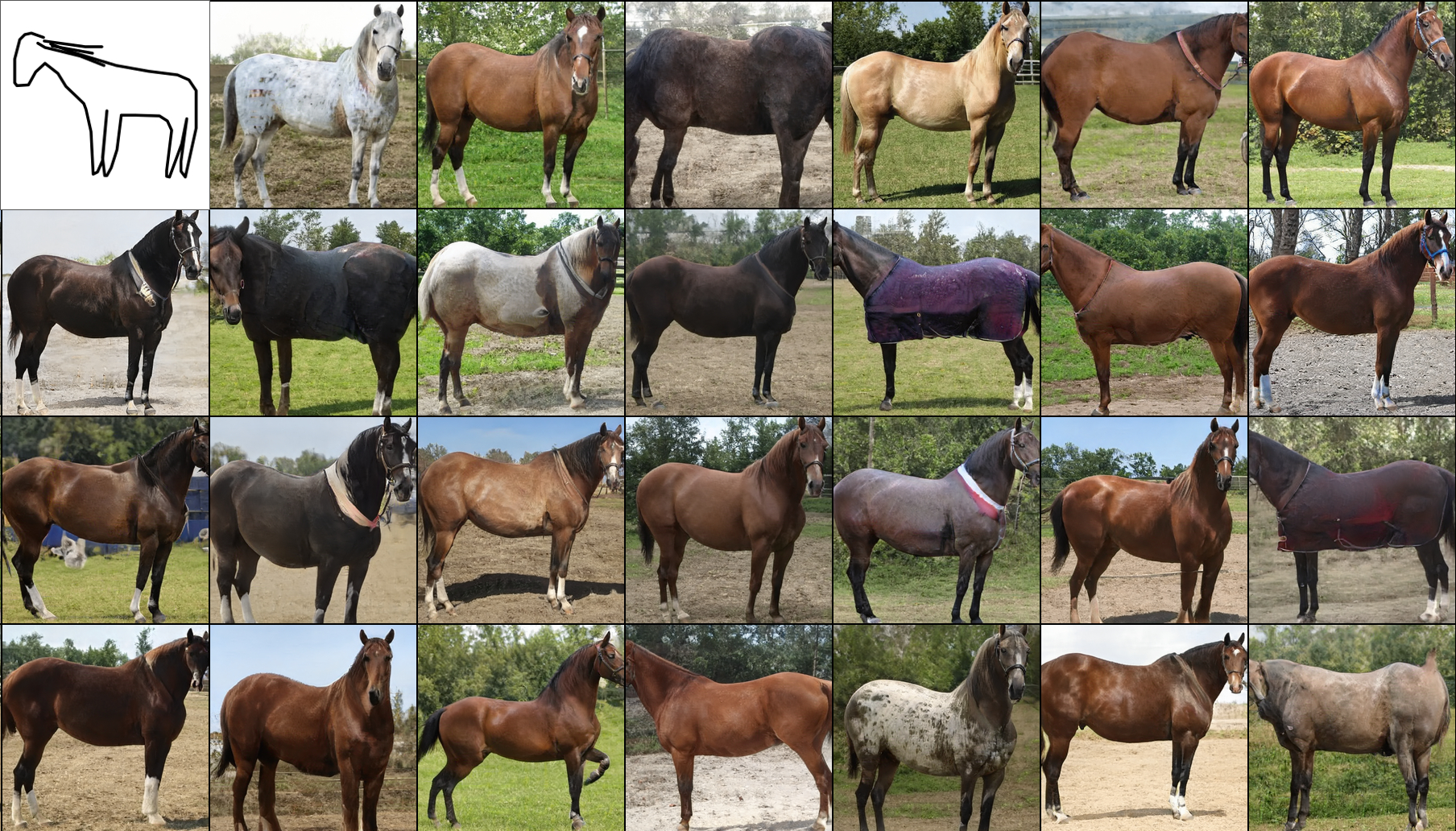}
  \caption{\textbf{Uncurated results }}
  \label{fig:uncuratedhorse0}
\end{figure*}

\begin{figure*}[h]
  \centering
  \includegraphics[width=\linewidth]{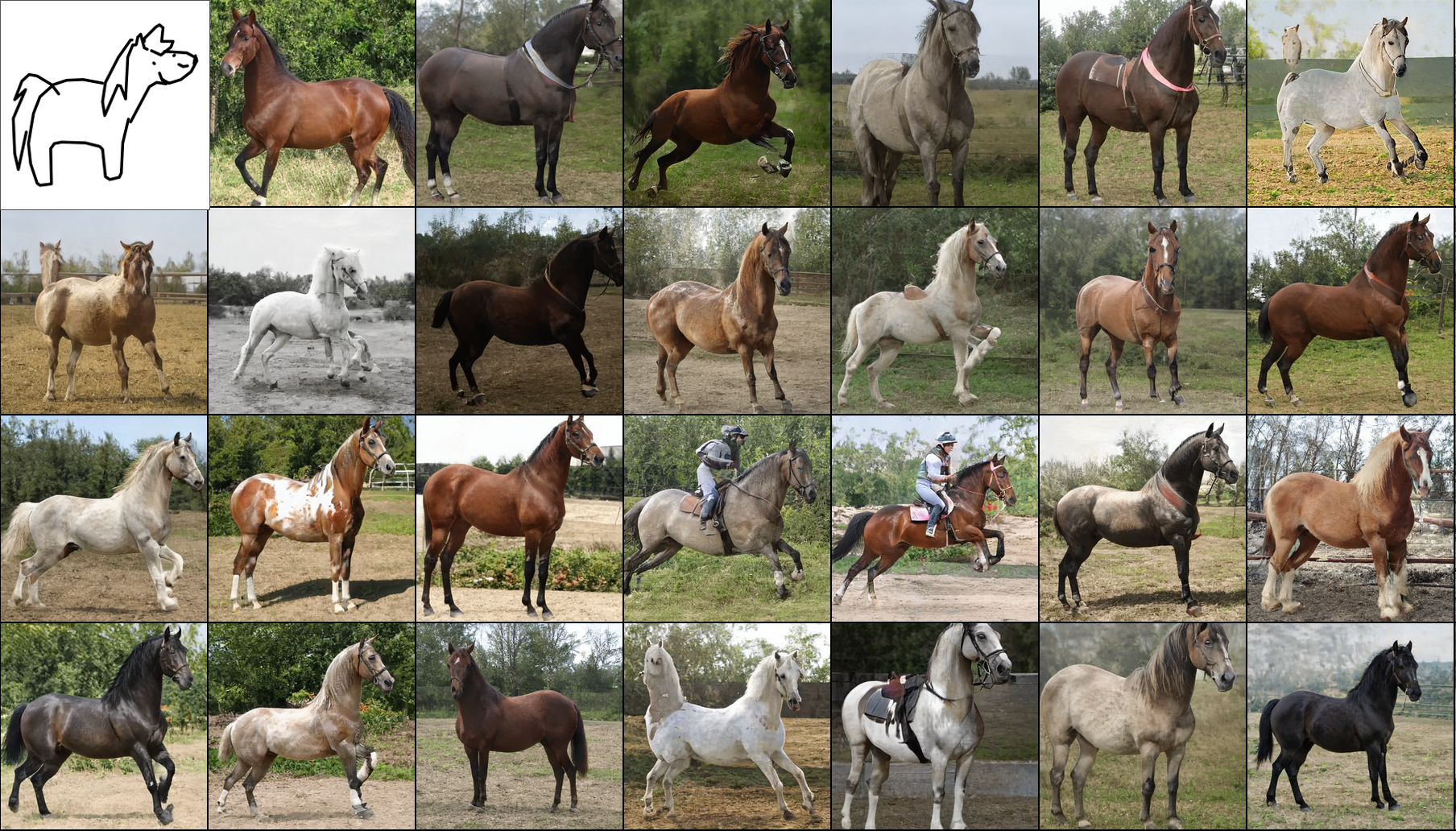}
  \caption{\textbf{Uncurated results }}
  \label{fig:uncuratedhorse1}
\end{figure*}

\begin{figure*}[h]
  \centering
  \includegraphics[width=\linewidth]{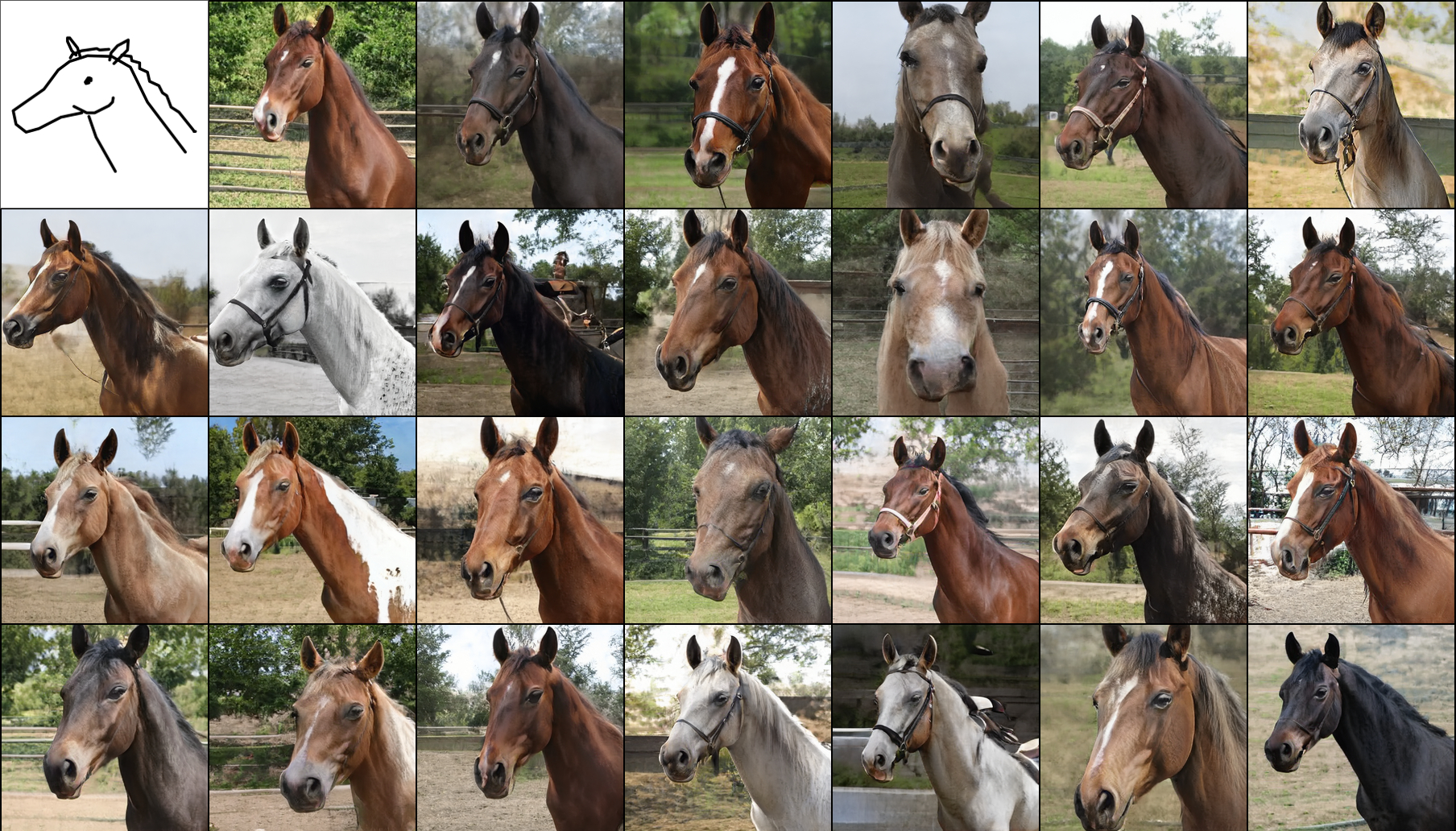}
  \caption{\textbf{Uncurated results }}
  \label{fig:uncuratedhorse2}
\end{figure*}

\begin{figure*}[h]
  \centering
  \includegraphics[width=\linewidth]{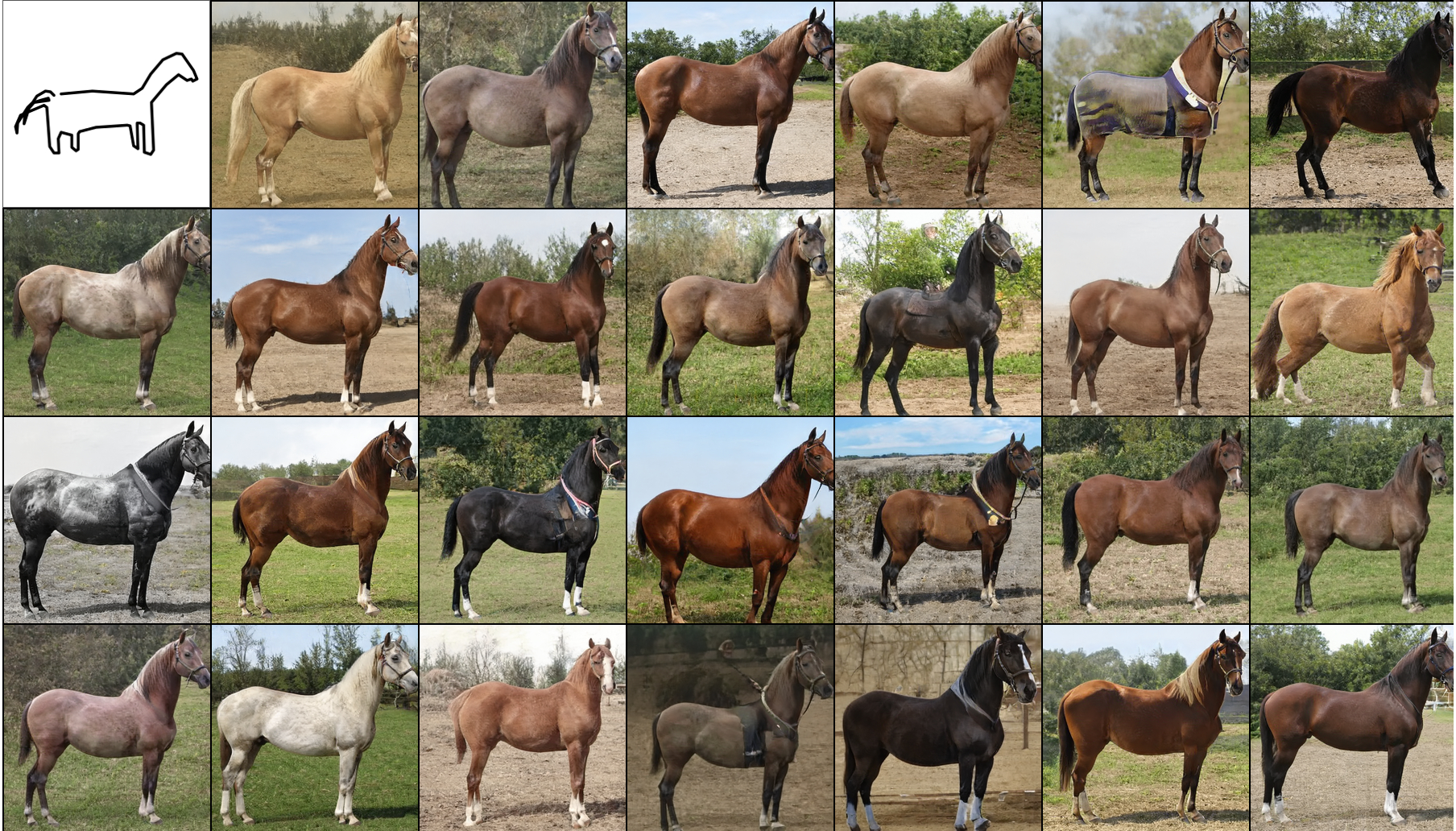}
  \caption{\textbf{Uncurated results }}
  \label{fig:uncuratedhorse3}
\end{figure*}

\begin{figure*}[h]
  \centering
  \includegraphics[width=\linewidth]{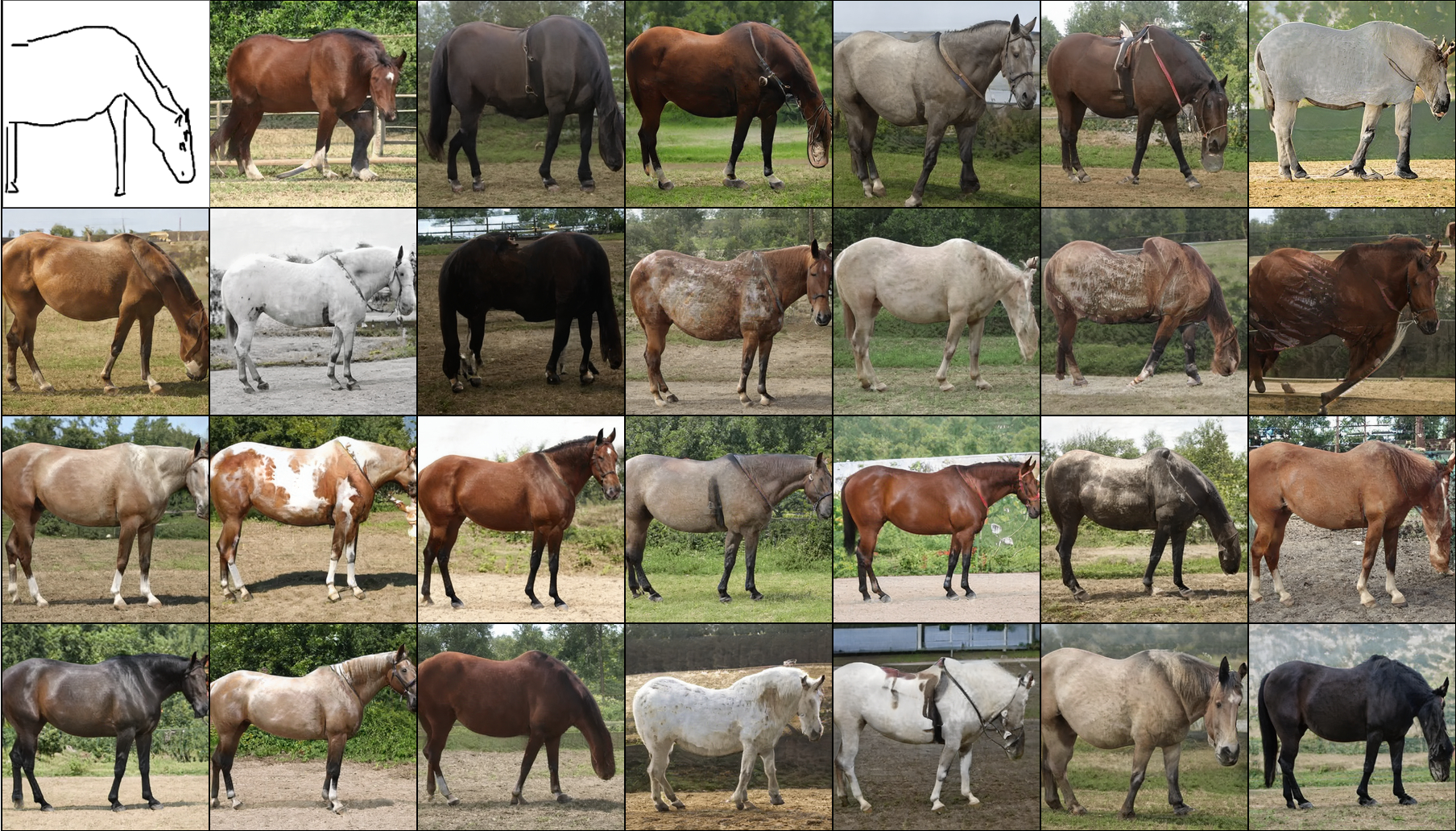}
  \caption{\textbf{Uncurated results }}
  \label{fig:uncuratedhorsesketchy1}
\end{figure*}

\begin{figure*}[h]
  \centering
  \includegraphics[width=\linewidth]{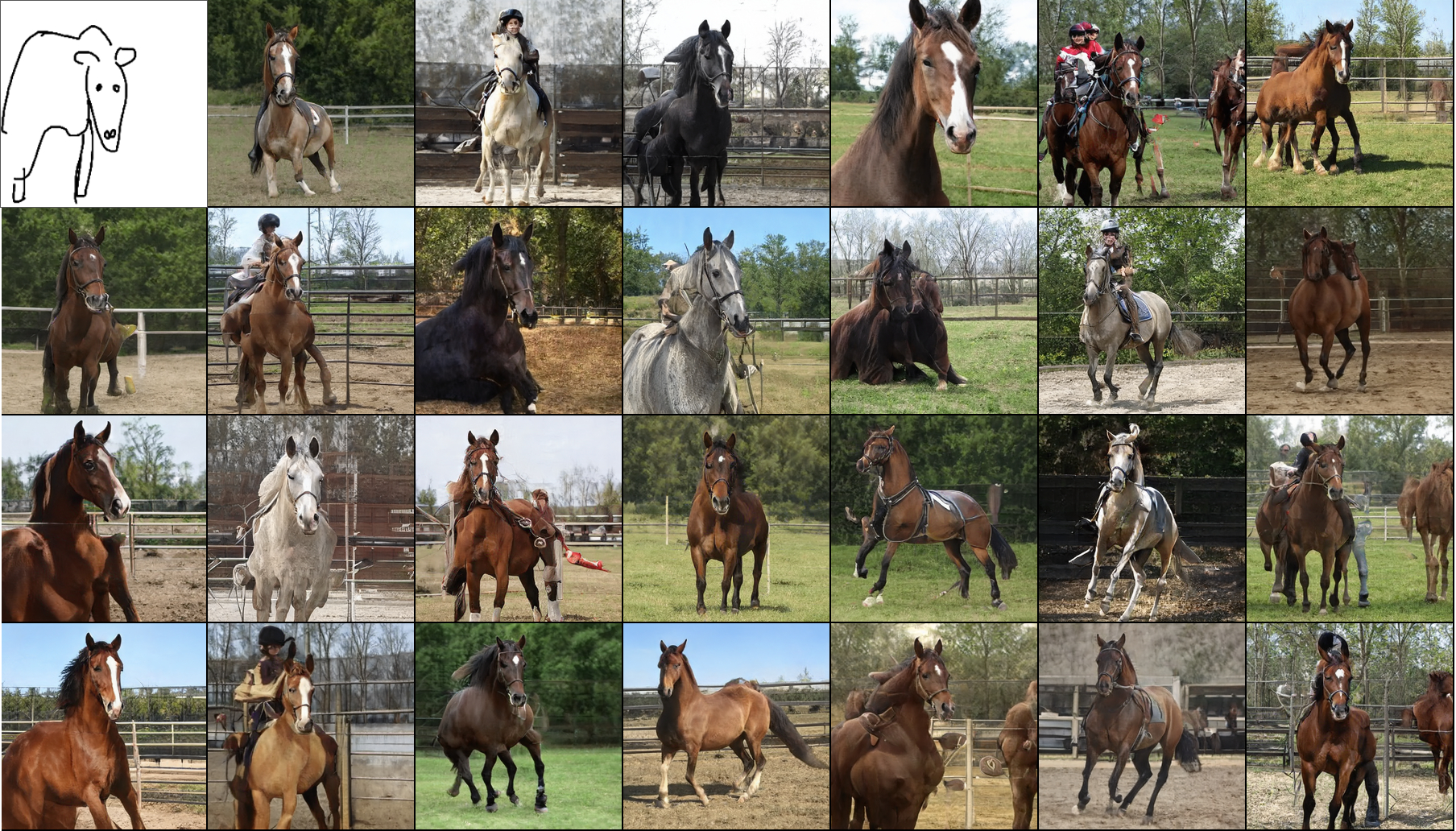}
  \caption{\textbf{Uncurated results }}
  \label{fig:uncuratedhorsesketchy2}
\end{figure*}

\textbf{Additional qualitative results. }
For all models successfully trained with hand sketches as listed in \ref{fig:handsketch}, we show more uncurated samples for each. Figure \ref{fig:uncuratedcat0} to \ref{fig:uncuratedcatsketchy2} are controlled samples based on the StyleGAN cat model pre-trained on LSUN-cat. Figure \ref{fig:uncuratedhorse0} to \ref{fig:uncuratedhorsesketchy2} are controlled samples based on the StyleGAN horse model.

\end{document}